\documentclass{article}
\usepackage{iclr2027_conference,times}
\iclrfinalcopy % Show the authors and suppress ICLR review line numbers.
\usepackage[utf8]{inputenc}
\usepackage[T1]{fontenc}
\usepackage{amsmath,amssymb,amsthm}
\usepackage{booktabs,multirow,tabularx,array}
\usepackage{colortbl}
\usepackage{microtype,graphicx,float}
\usepackage{tikz}
\usetikzlibrary{arrows.meta,positioning,calc}
\usepackage{pgfplots}
\pgfplotsset{compat=1.16}
\usepackage{hyperref,url}
\usepackage{amsmath,amsfonts,bm}

\def\eqref#1{equation~\ref{#1}}
\def\1{\bm{1}}

\DeclareMathAlphabet{\mathsfit}{\encodingdefault}{\sfdefault}{m}{sl}
\SetMathAlphabet{\mathsfit}{bold}{\encodingdefault}{\sfdefault}{bx}{n}

\definecolor{RGBlue}{HTML}{0072B2}
\definecolor{RGOrange}{HTML}{D55E00}
\definecolor{RGGreen}{HTML}{009E73}
\newcommand{\method}{\textsc{Fold2Reason}}
\newcommand{\base}{\textsc{Base}}
\newcommand{\pp}{\,\mathrm{pp}}

\title{Does Learning Protein Folding Generalize \\ to Broader Reasoning?}
\author{%
\begin{tabular}{@{}c@{}}
{\normalfont\normalsize\bfseries
Yong Liu\textsuperscript{1}\enspace
Zhanpeng Shi\textsuperscript{2,3}\enspace
Yizhou Dang\textsuperscript{4}\enspace
Zhongyue Zhang\textsuperscript{1}}\\[0.15em]
{\normalfont\normalsize\bfseries
Xiaoliang Shi\textsuperscript{1}\enspace
Zhijian Wei\textsuperscript{1}\enspace
Shuangjia Zheng\textsuperscript{1}}\\[0.3em]
{\normalfont\footnotesize
\textsuperscript{1}Shanghai Jiao Tong University\enspace
\textsuperscript{2}Fudan University}\\[0.1em]
{\normalfont\footnotesize
\textsuperscript{3}Shanghai Innovation Institute\enspace
\textsuperscript{4}Northeastern University}\\[0.1em]
{\normalfont\footnotesize\texttt{shuangjia.zheng@sjtu.edu.cn}}
\end{tabular}%
}
\date{}
\hypersetup{
  pdftitle={Fold2Reason: Learning Protein Folding for General Reasoning},
  pdfauthor={Yong Liu, Zhanpeng Shi, Yizhou Dang, Zhongyue Zhang, Xiaoliang Shi, Zhijian Wei, Shuangjia Zheng},
  pdfkeywords={Large language models, post-training, protein folding, geometry supervision}
}

\newcommand{\gentelheader}{%
  \includegraphics[height=0.27cm,trim=67 88 71 90,clip]{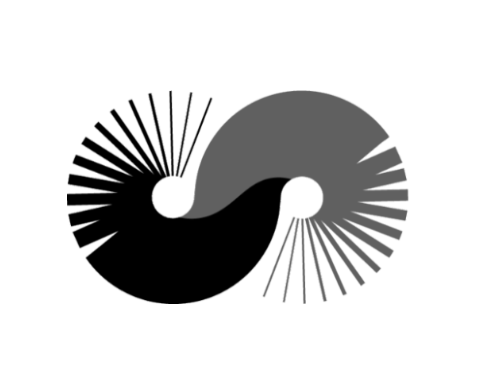}%
  \hspace{0.5em}\raisebox{0.06cm}{\scriptsize\sffamily\color{gray}GENTEL Lab}}

\begin{document}

\maketitle
\fancyhead[L]{\gentelheader}
\fancyhead[R]{}
\setlength{\headheight}{24pt}
\renewcommand{\headrulewidth}{0.3pt}

\begin{abstract}

Large language models rely heavily on human text, which often conveys surface answers rather than the spatial and structural logic behind them. Protein folding is a natural testbed, because one solved structure yields thousands of exactly checkable spatial and topological statements. We ask: can learning to fold proteins teach general models reusable reasoning capabilities?  To answer this, we build \textbf{FoldingCorpus}, a protein-derived question--answer dataset, and \textbf{Fold2Reason}, a recipe that post-trains on it through two complementary signals: discrete structural answers predicted via the model's native language head, and continuous 3D geometry decoded from the same shared representations. On FoldBench, \textbf{Fold2Reason} achieves structure prediction scores 2.7 to 3.5 times those of Qwen3.5-9B. Beyond protein structure prediction, it improves performance on all 10 benchmarks spanning spatial, graph, scientific, and general reasoning, raising macro-average accuracy from 45.09\% to 48.33\% (+3.23 pp), with positive gains on all 10 benchmarks, while matched controls built from random, synthetic, and shuffled structure yield substantially smaller or negative gains. Our work shows that non-linguistic, structure-dense scientific data can systematically improve broad reasoning in language models, making a solved scientific problem a practical source of post-training supervision.

\par\smallskip\noindent\textbf{Code:} \url{https://github.com/GENTEL-lab/Fold2Reason}

\end{abstract}

% (a) GPT Claude Qwen3.5-9B，要叫LM-Folding performance
% (b) 备注（Qwen3.5-9B）

\begin{figure}[H]
  \centering
  \includegraphics[width=0.999\linewidth]{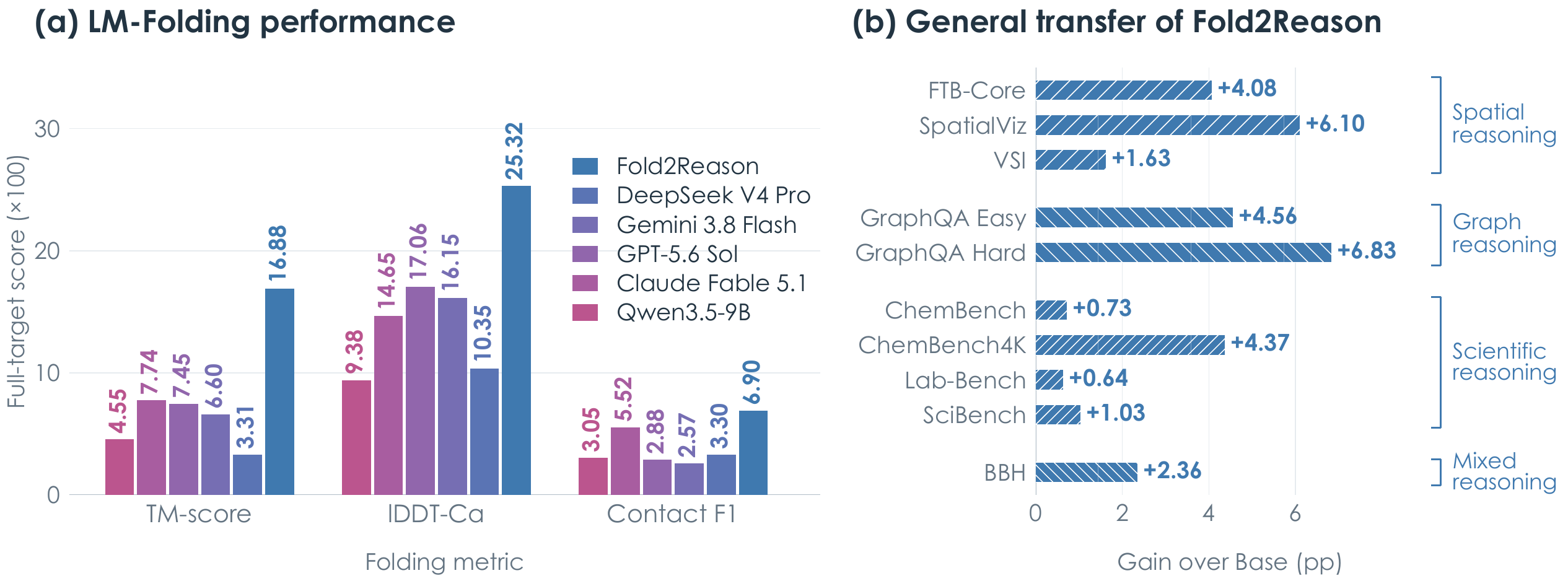}
\caption{\textbf{Folding performance and transfer to general reasoning.}
  (a) Full-target folding scores on all 334 FoldBench proteins \citep{xu2025foldbench}, for
  \method{} and for general-purpose language models prompted to emit C$\alpha$
  coordinates directly.
  (b) Gains of \method{} over its base model on the ten General-10 benchmarks, in
  percentage points, grouped into the four reasoning categories used throughout the
  paper.}
  \label{fig:overview}
\end{figure}

\section{Introduction}

Finite human-written text and code motivate alternative supervision for language
models \citep{raffel2020t5,gao2020pile,soldaini2024dolma,villalobos2024data,muennighoff2023dataconstrained}.
Instruction tuning, synthetic logic, and code training can improve behavior beyond
their source domains
\citep{ouyang2022instructgpt,chung2024flan,wang2023selfinstruct,morishita2024logic,ma2023codestage}.
Yet specialized training can also leave broader behavior unchanged or worse
\citep{huan2025mathtransfer,liu2026generalizationtax}. Which properties of supervision
support general transfer remains open.

We hypothesize that structurally constrained tasks can help a model acquire, or more
reliably invoke, computation reusable outside the training domain. We test behavioral
transfer without claiming to identify its underlying computation. Our source should
offer multiple structural targets, automatic answer verification, and scale; these
criteria do not imply that more labels necessarily yield more transfer.

Protein folding provides this setting. Advances in structure prediction
\citep{jumper2021alphafold,abramson2024alphafold3,baek2021rosettafold,ahdritz2024openfold},
geometric learning \citep{fuchs2020se3transformer,satorras2021egnn}, and protein
language models \citep{rives2021biological,lin2023evolutionary,su2024saprot,hayes2025esm3}
demonstrate learnable sequence--structure relationships. Solved coordinates in the
Protein Data Bank \citep{wwpdb2019} yield deterministic contact, distance, orientation,
and coordinate targets without additional annotation, enabling a test of transfer
from scientific structures to non-protein tasks.

We construct \textbf{FoldingCorpus} by applying 12 deterministic structural operators
to the OpenFold monomer short-protein collection \citep{ahdritz2024openfold}, producing question--answer examples
verified against source coordinates. FoldingCorpus names this dataset and its discrete
answer supervision; \method{} combines it with continuous geometry supervision.

\method{} post-trains Qwen3.5-9B \citep{qwen35modelcard} through a shared,
training-time workspace. FoldingCorpus answers supervise its native language head;
a frozen coordinate and distogram decoder supplies continuous geometry supervision
to the same LoRA-adapted representations \citep{hu2022lora}. External evaluation
retains only the adapted language model, removing protein inputs, the workspace,
and the geometry decoder (Figure~\ref{fig:overview}). This tests whether the learned
parameter updates transfer beyond the protein interface used to obtain them.

Across three independently trained adapters, \method{} raises the General-10 macro
from 45.09\% to 48.33\% ($+3.23\pp$), with positive dataset means across text, image,
and video tasks. FoldingCorpus-only gains extend across three Qwen3.5 scales
\citep{qwen35modelcard2b,qwen35modelcard4b,qwen35modelcard} and
InternVL3.5-8B \citep{wang2025internvl35}, while Gemma-4-12B-IT
\citep{gemmateam2026gemma4} is neutral. An independent 50--4,000-protein
study peaks at $+3.70\pp$ with 2,000 proteins under a fixed-epoch schedule that
increases coverage and optimization steps together. Component ablations locate most
transfer in FoldingCorpus supervision ($+2.93\pp$ without Geometry); Geometry adds
$0.30\pp$ overall, with its increments concentrated in spatial evaluations and local
structural readouts. These results support behavioral transfer with model and task
dependence, rather than uniform improvement in reasoning.

Our main contributions are:
\begin{itemize}
  \item \textbf{Data infra.} We construct a new post-training dataset
  , \textbf{FoldingCorpus}, from existing protein structures. Its core contains 1,200 proteins and 14,400
  question--answer records across 12 structural operators, with 1,000 training,
  100 development, and 100 frozen-test proteins. Cluster-disjoint core partitions
  and independently recomputed answers make the supervision auditable.
  FoldingCorpus turns scientific coordinates into a concrete data resource for
  studying general reasoning transfer.

  \item \textbf{Method.} We convert known structures into two exactly checkable
  training signals and route them through a single shared interface: discrete
  FoldingCorpus answers supervise the native language head, while a frozen coordinate
  decoder constrains the same LoRA-adapted residue states. Downstream evaluation uses
  the adapted base model alone.
  
  \item \textbf{Findings.} Post-training on protein structure improves a general
  language model on reasoning tasks that contain no proteins, no structural inputs, and
  no specialized modules. Across three seeds and 10 general-purpose benchmarks,
  \method{} raises the macro-average from 45.09\% to 48.33\% ($+3.23\pp$), with positive
  mean changes on all 10 datasets.
\end{itemize}

\section{Related Work}

\paragraph{Protein folding and its alignment with language models.}
Supervised geometric pipelines made structure prediction reliable
\citep{jumper2021alphafold,baek2021rosettafold,ahdritz2024openfold,
abramson2024alphafold3}, and a recent review reports that predicted coordinates now
serve as working models across drug discovery, enzyme engineering, and disease biology
\citep{yin2026structureprediction}. Geometric generative models also synthesize
protein backbones using flow matching \citep{bose2024foldflow}. A parallel line moves
structure into language models: protein language models fold directly or absorb
structural states into their vocabulary
\citep{lin2023evolutionary,su2024saprot,hayes2025esm3}, and recent work aligns a
sequence model with a structural graph encoder \citep{chen2025structurealigned}.
Protein encoders are also connected to general-purpose LLMs for protein understanding
\citep{xiao2025proteingpt,shu2024aligning,xiao2025proteinllmsurvey}, while 3D-MoLM
aligns molecular structure with text for molecule--text tasks
\citep{li2024threedmolm}. These alignment efforts evaluate molecular or protein
capabilities. We instead use protein structure as a training signal and measure
downstream reasoning after the protein view, workspace, and geometry decoder have
been detached.

\paragraph{What training data builds general capability.}
A complementary literature asks which data makes a model more capable. Corpus work
documented composition \citep{raffel2020t5,gao2020pile,soldaini2024dolma} before
attention turned to the ceiling of human-written text
\citep{villalobos2024data,muennighoff2023dataconstrained}. Recent work therefore
engineers data rather than collecting it, by targeting pretraining selection at
downstream tasks \citep{mizrahi2025betr}, synthesizing corpora
\citep{yang2025synthetic,morishita2024logic}, and training against verifiable answers
\citep{lambert2025tulu3,deepseek2025r1,ma2025generalreasoner}. How far such training
travels is now measured directly, with mixed results
\citep{ma2023codestage,huan2025mathtransfer,chu2025sftrl,liu2026generalizationtax}.
Complementary studies identify distribution mismatch in offline self-correction
training \citep{kumar2025score} and analyze how finetuning changes predictions on
other examples \citep{ren2025learningdynamics}. A recent survey organizes the resulting
data-centric design space \citep{liang2026datacentric}. Program-generated logic and
verifiable-reward training already provide checkable supervision
\citep{morishita2024logic,lambert2025tulu3,deepseek2025r1}. FoldingCorpus draws its
answer targets from three-dimensional protein coordinates, extending these sources of
structured supervision to a scientific structural archive.

\section{Method}
\label{sec:method}

\subsection{Training targets: known answers, hidden evidence}

For a protein of length $L$, let $x$ contain the amino-acid sequence and optional MSA
or template evidence, $Y\in\mathbb{R}^{L\times4\times3}$ its known backbone
coordinates, and $m$ the residue-validity mask. Coordinates generate targets and
losses and remain outside the model input. Qwen processes a prompt with one
marker per residue, producing
\begin{equation}
H=f_{\theta}(x)_{\mathrm{res}}\in\mathbb{R}^{L\times4096},
\end{equation}
where the base weights are frozen and $\theta$ includes trainable LoRA parameters.

Each protein yields one packed set of 12 questions. Programs over $Y$ create contact,
distance comparison, segment orientation, center proximity, local direction,
chirality, multi-constraint, and global-summary labels. Nine answers are binary, two
are three-way, and one is a 32-way textual summary match; every option is one
vocabulary token. For example, the program computes the answer to ``do residues 23
and 147 contact?'' from $Y$, and Qwen predicts that target from the protein view and
learned states. Appendix~\ref{app:foldingcorpus_operators} gives the mathematical
definition, threshold, sampling rule, and observed training-label count for every
operator.

We construct these packs once, independently recompute the answers, and retain
the numerical evidence in an audit record rather than the prompt. Independent
hash salts determine label sampling, option order, and question order; the frozen
packs are reused across epochs. The 32-way question chooses among one target
summary and 31 hard-negative summaries. It remains ordinary answer-token
supervision, distinct from the separate retrieval-projection objective, which is
disabled in the canonical recipe.

\begin{figure}[t]
  \centering
  \includegraphics[width=\linewidth]{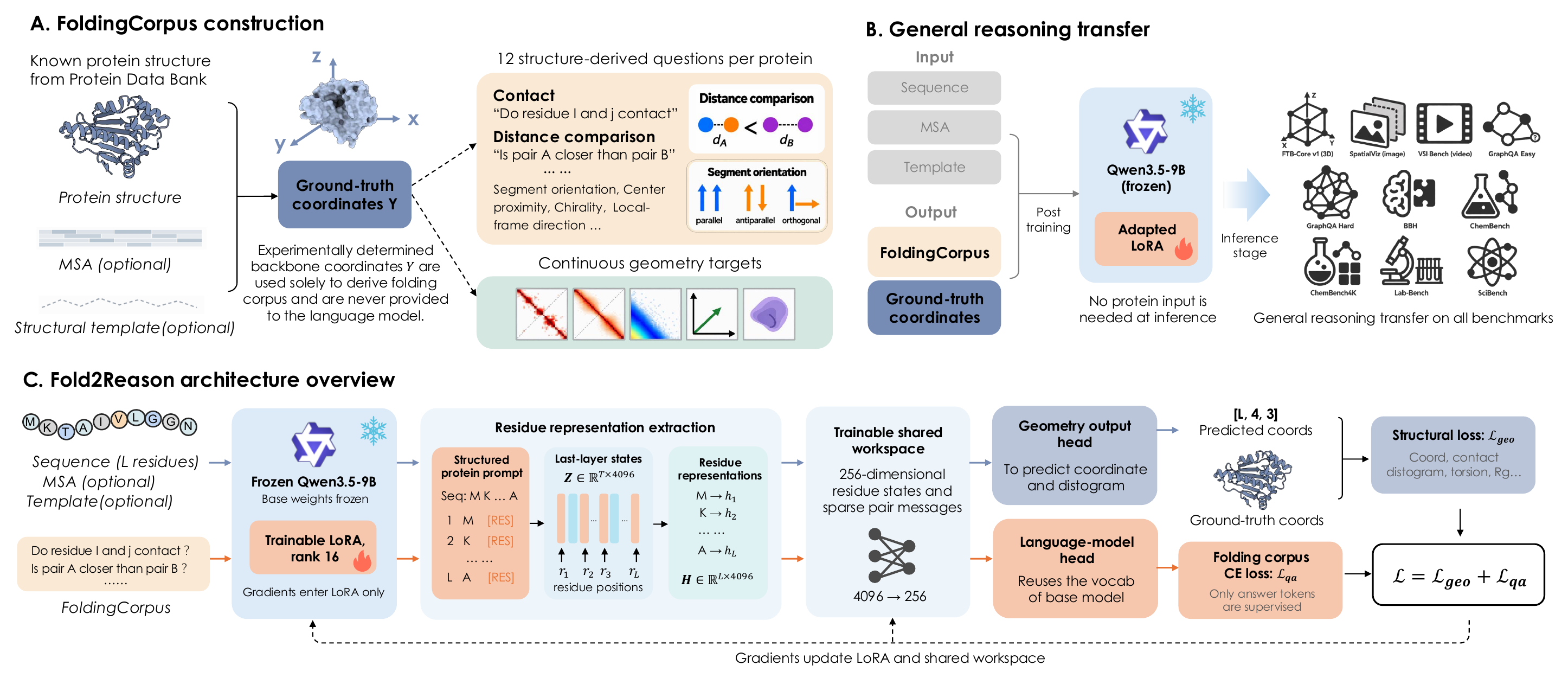}
  \caption{\textbf{The \method{} pipeline from protein structures to general
  reasoning transfer.} (a) Known protein structures generate deterministic FoldingCorpus labels and continuous geometry targets; ground-truth coordinates are used only for supervision. (b) The adapted model transfers to general reasoning benchmarks without protein inputs at inference. (c) A frozen language-model backbone with trainable LoRA and a shared workspace feeds FoldingCorpus and geometry readouts during
  post-training.}
  \label{fig:architecture}
\end{figure}

\subsection{One workspace, two readouts}

The workspace $W_{\phi}$ reduces marker states to 256 dimensions, exchanges messages
over sequence-local pairs and sampled long-range pairs, and returns two readouts:
\begin{equation}
(E,M)=W_{\phi}(H),\qquad
E\in\mathbb{R}^{L\times4096},\quad M\in\mathbb{R}^{16\times4096}.
\end{equation}
$E$ is residue aligned. Sixteen learned queries pool the residue set and project it
back to LM width, giving evidence tokens $M$. ``Workspace'' refers only to these
training-time residue and pooled tensors.

Pairs connect sequence offsets 1--4 and evenly spaced longer-range indices, with
at most 2,048 pairs per protein; target contacts do not select the edges.
Symmetric features combine absolute differences and elementwise products of the
reduced states. Messages are averaged at incident residues, followed by residual
updates. Learned-query pooling forms $M$, providing a fixed-size prefix while $E$
preserves residue-level correspondence \citep{jaegle2021perceiver,li2021prefixtuning}.

The FoldingCorpus path prepends $M$ to the packed question sequence $q$. Prefix and
prompt positions are ignored; only the 12 answer tokens and EOS are supervised, giving
the answer loss $\mathcal{L}_{\mathrm{qa}}$:
\begin{equation}
\mathcal{L}_{\mathrm{qa}}=-\frac{1}{|\mathcal{S}|}
\sum_{t\in\mathcal{S}}\log p_{\theta,\phi}(y_t\mid M,q,y_{<t}).
\end{equation}
The geometry path passes $E$ to a frozen coordinate and distogram decoder $g_{\psi}$.
Its loss combines coordinate, pair-distance, contact, distogram, local-frame, torsion,
and radius-of-gyration terms:
\begin{align}
\mathcal{L}_{\mathrm{geo}}={}&\mathcal{L}_{\mathrm{coord}}+
\mathcal{L}_{\mathrm{pair}}+\mathcal{L}_{\mathrm{contact}}+
\mathcal{L}_{\mathrm{dist}}+\mathcal{L}_{\mathrm{local}}+\mathcal{L}_{\mathrm{torsion}}+
\mathcal{L}_{R_g}.
\end{align}
Freezing $\psi$ prevents a new coordinate head from absorbing the objective; gradients
must change the shared workspace and LoRA. The canonical objective is
$\mathcal{L}=\mathcal{L}_{\mathrm{qa}}+\mathcal{L}_{\mathrm{geo}}$. One FoldingCorpus
target is a 32-way textual classification task and is trained through the same native
language-model head as the other FoldingCorpus answers. Appendix~\ref{app:geometry_decoder}
documents the provenance, training data, and freeze checks for $g_{\psi}$.

Each training example uses two forwards through the same LoRA-adapted Qwen: the
protein forward produces $H$, and the answer forward consumes the concatenation
of $M$ with the embedded question--answer sequence. We retain the computation
graph between them, so answer loss updates both the answering parameters and the
protein-to-workspace path. Freezing the decoder means excluding $\psi$ from the
optimizer, not detaching $E$; geometry gradients therefore still reach $\phi$ and
the shared LoRA parameters. This limits adaptation of the readout itself, without
equating structural decodability with a particular reasoning mechanism
\citep{hewitt2019probes,ravichander2021probing}.

Only LoRA and the active workspace parameters are optimized. In the canonical
run, four workers accumulate two one-protein microsteps, giving eight proteins per
update. We average answer CE over the 12 labels and EOS, combine it with geometry
loss, and clip the accumulated gradient norm to 1.0 before each AdamW update
\citep{loshchilov2019adamw}.
At transfer evaluation, we discard $W_{\phi}$ and $g_{\psi}$ and apply only the
learned LoRA adapter to the model's native benchmark interface. Thus, any measured
transfer must reside in the adapted model rather than the protein-specific modules.

\section{Experimental Design}

\paragraph{Training.}
We train Qwen3.5-9B \citep{qwen35modelcard} on 1,000 proteins: 336 sequence-only,
332 sequence+MSA, and 332 sequence+MSA+template views. Lengths range from 29 to 199. The resulting 12,000 FoldingCorpus records are packed into one example per protein. LoRA is applied to audited attention and feed-forward projections (rank 16, alpha 32, dropout 0.05; 43.28M parameters); the workspace has 3.90M parameters and the frozen decoder 3.31M. Every run uses three epochs, 375 optimizer steps, four A800-80G GPUs, and seeds 20260729, 20260803, and 20260804. All checkpoints are evaluated at epoch 3/step 375, without benchmark-based selection. Appendix~\ref{app:training} gives the complete contract, and Appendix~\ref{app:geometry_decoder} gives the frozen decoder protocol.

\paragraph{Evaluation.}
General-10 is the unweighted macro over FTB-Core, SpatialViz, VSI, GraphQA Easy and
Hard, BBH, ChemBench, ChemBench4K, Lab-Bench, and SciBench:
\begin{equation}
\Delta_{\mathrm{G10}}=\frac{1}{10}\sum_{d=1}^{10}
\left[s_d(\mathrm{adapter})-s_d(\base)\right].
\end{equation}
FTB-Core is our internally constructed text benchmark for transferable 3D reasoning.
It contains 12,000 fixed questions covering spatial primitives, constraint
satisfaction, packing and clearance, SE(3) transformations and symmetry, and noisy
evidence fusion. SpatialViz evaluates image-grounded spatial reasoning
\citep{wang2025spatialviz}, and VSI evaluates video-grounded spatial understanding
\citep{yang2025thinking}. GraphQA Easy and Hard test reasoning over graph structure
\citep{fatemi2023graph}. BBH covers diverse hard reasoning tasks
\citep{suzgun2022bbh}. ChemBench \citep{mirza2025chembench} and ChemBench4K from
ChemLLM \citep{zhang2024chemllm} are separately sourced chemistry question-answering
benchmarks. Lab-Bench targets laboratory and scientific workflow reasoning
\citep{laurent2024labbench}, and SciBench tests scientific problem solving
\citep{wang2024scibench}. All 76,725 examples are included. FTB-Core, SpatialViz, and VSI enter the macro as
three of the ten datasets and carry no extra weight. FoldBench334 uses the monomer
targets from FoldBench \citep{xu2025foldbench}, drawn from the Protein Data Bank
\citep{wwpdb2019}; we report TM-score \citep{zhang2004tmscore}, lDDT-C$\alpha$
\citep{mariani2013lddt}, contact F1, and C$\alpha$ distance MAE.
Figure~\ref{fig:overview}(a) additionally compares the full folding system with
direct coordinate generation by general-purpose language models, including the
unadapted Qwen3.5-9B backbone. This comparison averages all 334 targets, including
incomplete predictions, using the full-target scoring rules in
Appendix~\ref{app:native_folding}.
Appendices~\ref{app:ftb_core} and~\ref{app:evaluation_protocol} document the fixed
FTB-Core generator and the prompt, decoding, media-sampling, parser, and invalid-answer
contract for every benchmark. External9 applies the same unweighted macro to the nine
public benchmarks and isolates transfer beyond the internally generated FTB-Core.

The unit of replication is an independently trained adapter. We report the mean and
sample SD across three training seeds.

\paragraph{Ablations and controls.}
The ablation arms follow the naming in Table~\ref{tab:components}.
FoldingCorpus-only denotes training on the dataset's answer loss with no Geometry
objective. It appears in two configurations: the w/o Geometry arm of
Table~\ref{tab:components}, which keeps the workspace, and the Pure-LoRA setting used
for the matched controls and the model-family study, which does not.
The w/o FoldingCorpus arm removes FoldingCorpus CE while retaining protein inputs,
the Geometry objective, and the historical retrieval auxiliary; we therefore
analyze it as a geometry-dominant historical arm.
Fold2Reason-full combines FoldingCorpus CE and Geometry; appendix tables and figures
label this arm Full RG. The component
ablations use the same base model, training
endpoint, seeds, and General-10 evaluation protocol as Fold2Reason-full. We use these
ablations to localize which loss path drives external transfer and which path improves
structural readout.

\section{Results}

\subsection{General-10 transfer and matched FoldingCorpus-format controls}

\method{} improves General-10 from 45.09\% to 48.33\%, a change of
$3.23\pp$ (SD $0.15\pp$). Seed changes are $3.19$,
$3.12$, and $3.40\pp$.
The largest mean changes occur on GraphQA Hard ($+6.83\pp$), SpatialViz
($+6.10\pp$), GraphQA Easy ($+4.56\pp$), and ChemBench4K ($+4.37\pp$), and the
text-only seven-dataset macro rises by $2.93\pp$. Removing GraphQA Easy and Hard leaves
an eight-dataset macro gain of $2.62\pp$ and a five-dataset text-only gain of
$1.83\pp$, and excluding FTB-Core leaves an External9 gain of $+3.14\pp$. The aggregate gain is therefore distributed across
modalities and benchmark families, with positive mean changes on all 10 datasets.

\begin{table}[H]
\centering
\caption{
Dataset-level General-10 results for matched FoldingCorpus-format controls and the full
\method{} recipe. Values are mean accuracies in percent over three seeds; parentheses
report absolute change from \base{} in percentage points. Categories follow
Figure~\ref{fig:overview}(b). Green/red cells indicate gains/declines; stronger shading
indicates larger absolute changes on a shared scale. The three controls share the same
LoRA recipe, supervised-token budget, seeds, and full General-10 evaluation.
}
\label{tab:protein_specificity_controls}

\scriptsize
\setlength{\tabcolsep}{1.3pt}
\renewcommand{\arraystretch}{1.22}

\begin{tabular*}{\linewidth}{@{\extracolsep{\fill}}ccccccc@{}}
\toprule
\multirow{2}{*}{\textbf{Category}}
& \multirow{2}{*}{\textbf{Benchmark}}
& \multirow{2}{*}{\textbf{\base{}}}
& \multicolumn{4}{c}{\textbf{Post-training}} \\
\cmidrule(lr){4-7}
& &
& \textbf{Hidden Geom.}
& \textbf{Format Copy}
& \textbf{Fixed Shuffle}
& \textbf{\method{}} \\
\midrule

\multirow{3}{*}{\shortstack[c]{Spatial\\reasoning}}
& FTB-Core
& 36.18
& \cellcolor[HTML]{C6E1BB}37.25 (+1.07)
& \cellcolor[HTML]{FBECEE}36.04 (-0.14)
& \cellcolor[HTML]{ECB9BD}35.41 (-0.77)
& \cellcolor[HTML]{A0CD8D}\textbf{40.26 (+4.08)} \\

& SpatialViz
& 26.61
& \cellcolor[HTML]{C96A73}23.16 (-3.45)
& \cellcolor[HTML]{B8DAAA}28.56 (+1.95)
& \cellcolor[HTML]{EAB5B9}25.76 (-0.85)
& \cellcolor[HTML]{8EC478}\textbf{32.71 (+6.10)} \\

& VSI Bench
& 58.53
& \cellcolor[HTML]{CEE6C5}59.22 (+0.69)
& \cellcolor[HTML]{D98B92}56.32 (-2.21)
& \cellcolor[HTML]{FCEFF0}58.42 (-0.11)
& \cellcolor[HTML]{BDDDB0}\textbf{60.16 (+1.63)} \\

\midrule

\multirow{2}{*}{\shortstack[c]{Graph\\reasoning}}
& GraphQA Easy
& 65.14
& \cellcolor[HTML]{BCDCAD}66.90 (+1.76)
& \cellcolor[HTML]{E6AAAF}63.95 (-1.19)
& \cellcolor[HTML]{F7DFE1}64.91 (-0.23)
& \cellcolor[HTML]{9BCB88}\textbf{69.70 (+4.56)} \\

& GraphQA Hard
& 32.65
& \cellcolor[HTML]{B5D9A7}34.83 (+2.18)
& \cellcolor[HTML]{C3DFB6}33.95 (+1.30)
& \cellcolor[HTML]{A8D298}35.88 (+3.23)
& \cellcolor[HTML]{88C171}\textbf{39.48 (+6.83)} \\

\midrule

\multirow{4}{*}{\shortstack[c]{Scientific\\reasoning}}
& ChemBench
& 66.76
& \cellcolor[HTML]{FCF0F1}66.64 (-0.12)
& \cellcolor[HTML]{CCE5C2}67.44 (+0.68)
& \cellcolor[HTML]{BF535D}62.85 (-3.91)
& \cellcolor[HTML]{CDE5C4}\textbf{67.49 (+0.73)} \\

& ChemBench4K
& 63.56
& \cellcolor[HTML]{B2D7A4}65.93 (+2.37)
& \cellcolor[HTML]{CCE5C2}64.35 (+0.79)
& \cellcolor[HTML]{DA8F96}61.44 (-2.12)
& \cellcolor[HTML]{9DCC8A}\textbf{67.93 (+4.37)} \\

& Lab-Bench
& 37.21
& \cellcolor[HTML]{C4E0B8}\textbf{38.43 (+1.22)}
& \cellcolor[HTML]{C7E2BC}38.24 (+1.03)
& \cellcolor[HTML]{EFC3C6}36.58 (-0.63)
& \cellcolor[HTML]{CFE6C6}37.86 (+0.64) \\

& SciBench
& 9.83
& \cellcolor[HTML]{B2D7A4}12.19 (+2.36)
& \cellcolor[HTML]{D0E7C7}10.40 (+0.57)
& \cellcolor[HTML]{B0D6A0}\textbf{12.36 (+2.53)}
& \cellcolor[HTML]{C7E2BC}10.86 (+1.03) \\

\midrule

\begin{tabular}[c]{@{}c@{}}Mixed reasoning\end{tabular}
& BBH
& 54.45
& \cellcolor[HTML]{C5E1BA}55.56 (+1.11)
& \cellcolor[HTML]{C45F68}50.79 (-3.66)
& \cellcolor[HTML]{DFEED9}54.61 (+0.16)
& \cellcolor[HTML]{B2D7A4}\textbf{56.81 (+2.36)} \\

\midrule

\multicolumn{2}{c}{\textbf{General-10 macro}}
& 45.09
& \cellcolor[HTML]{C9E3BF}46.01 (+0.92)
& \cellcolor[HTML]{FDF3F3}45.01 (-0.09)
& \cellcolor[HTML]{F6DCDE}44.82 (-0.27)
& \cellcolor[HTML]{A8D298}\textbf{48.33 (+3.23)} \\

\bottomrule
\end{tabular*}
\end{table}

Matched controls separate correct input--target correspondence from geometry-target and
answer-format alternatives. Three Pure-LoRA controls share the Qwen3.5-9B base model,
12-question packing, 375 optimizer steps, checkpoint selection, and three training seeds.
Hidden Geometry keeps the protein prompts but recomputes each pack's answers from one
unobserved random point structure; Format Copy supplies donor answer codes and trains the
model to copy them; Fixed Shuffle keeps the original prompts but assigns fixed donor
labels matched by operator and candidate vocabulary, detailed in Appendix~\ref{app:matched_control_details}.

Table~\ref{tab:protein_specificity_controls} shows that Hidden Geometry improves the
General-10 macro by $0.92\pp$, whereas Format Copy changes it by $-0.09\pp$
and Fixed Shuffle by $-0.27\pp$. Thus, input-independent but internally coherent
geometry targets retain some transfer, while copying valid answer codes or learning a
fixed input--label mismatch does not yield a consistent aggregate gain. Protein-derived
\method{} reaches $+3.23\pp$, more than twice the strongest control, so its benefit
cannot be explained by answer formatting or fixed shuffled supervision alone.
% BEGIN INLINED OUTPUT-VALIDITY AUDIT: core-results/output-format-audit-20260915/main_text.tex
The FTB-Core and SpatialViz gains persist on questions with valid outputs from
both models; Appendix~\ref{app:output_validity} reports the parsing audit,
task concentration, and remaining answer-selection-bias limitations.
% END INLINED OUTPUT-VALIDITY AUDIT: core-results/output-format-audit-20260915/main_text.tex

\subsection{FoldingCorpus supervision transfers across model scales and families}

\begin{figure}[H]
  \centering
  \includegraphics[width=\linewidth]{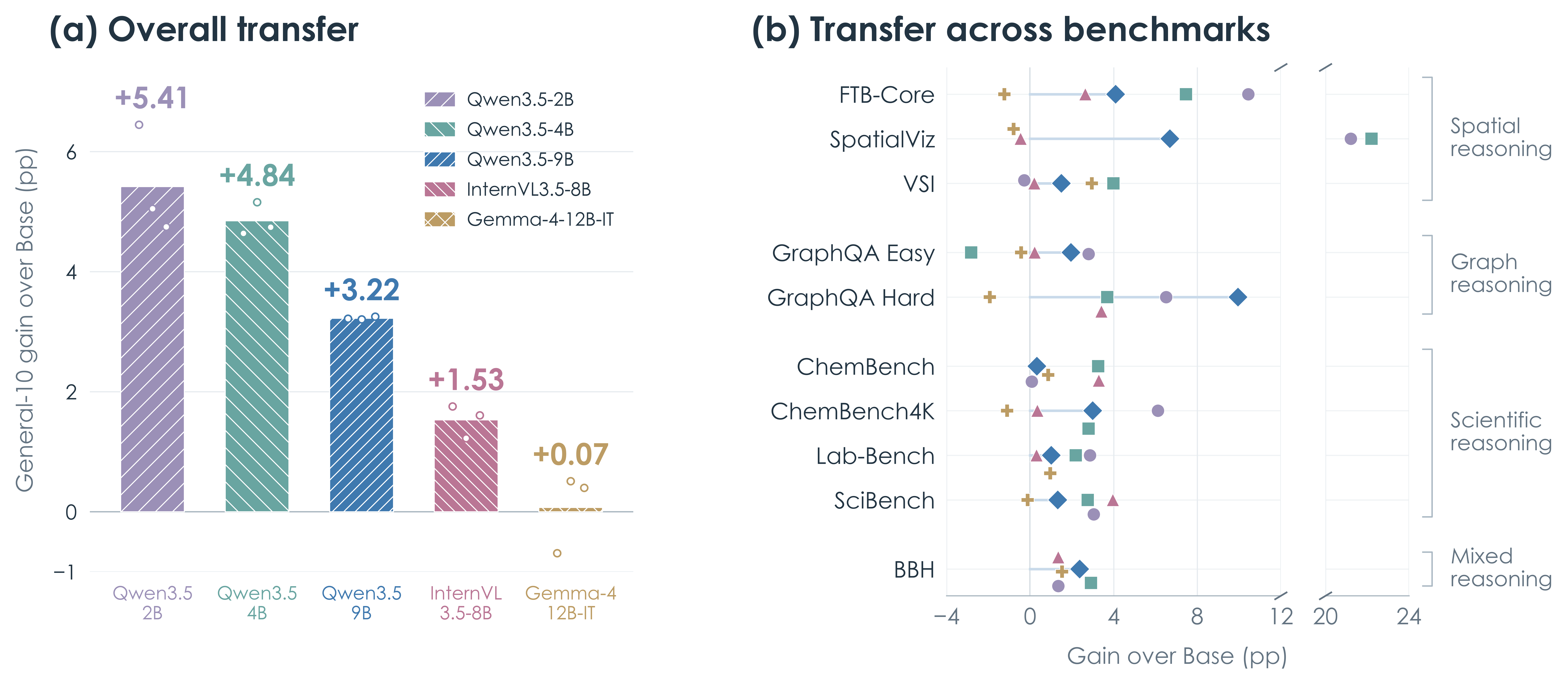}
  \caption{\textbf{FoldingCorpus supervision across model scales and families.}
  (a) General-10 change relative to each model's own base checkpoint. Bars show the
  three-seed mean and open circles show the individual seeds. (b) The same change
  resolved per benchmark, with one marker shape and colour per model and the
  categories of Figure~\ref{fig:overview}(b). The horizontal axis is broken to
  accommodate the large SpatialViz gains of the two smaller Qwen3.5 models.}
  \label{fig:model_family_transfer}
\end{figure}

We further test the model-level generality of FoldingCorpus supervision
with the same Pure-LoRA comparison on three Qwen3.5 scales and two independent
multimodal model families, InternVL3.5 and Gemma-4. Each setting is evaluated on the
full General-10 suite over three training seeds. Figure~\ref{fig:model_family_transfer}
shows that the improvement persists across all three Qwen3.5 scales and transfers to
InternVL3.5-8B. FoldingCorpus supervision raises General-10 by $5.41$, $4.84$, and
$3.22\pp$ for Qwen3.5-2B, 4B, and 9B, respectively. The InternVL result raises
General-10 by $1.53\pp$, with a positive change for every seed.

Figure~\ref{fig:model_family_transfer}(b) resolves these aggregates by benchmark and
shows that no single dataset carries them: all four reasoning categories contribute, and
the largest single effects are the SpatialViz gains of Qwen3.5-2B and 4B. Architecture
nonetheless shapes the magnitude: Gemma-4-12B-IT remains at its base level ($+0.07\pp$,
two positive and one negative seed), so the positive InternVL result establishes transfer
beyond Qwen while the neutral Gemma result identifies meaningful family dependence.

\subsection{General transfer under joint scaling of protein coverage and training step}

\begin{figure}[H]
  \centering
  \includegraphics[width=\linewidth]{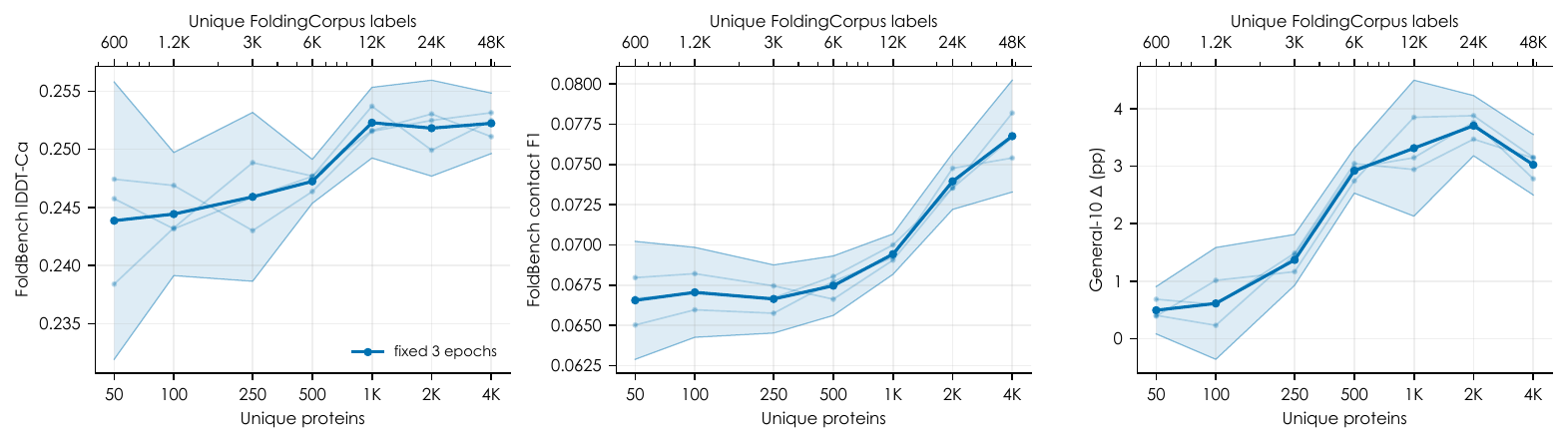}
  \caption{\textbf{Data scaling of structural readout and general transfer.}
  The three panels report FoldBench lDDT-C$\alpha$, FoldBench Contact F1, and
  General-10 change for an independent three-seed Qwen3.5-9B Full-RG scaling run over
  seven nested subsets spanning 50--4,000 proteins. Every subset is trained for three
  epochs. Thin curves show individual seeds, center curves show means, and transparent
  bands with boundary lines show three-seed 95\% $t$-intervals. The upper axis gives
  the corresponding number of FoldingCorpus records at 12 targets per protein.}
  \label{fig:data_scaling}
\end{figure}

We study seven nested training sets spanning 50--4,000 proteins, equivalent to
600--48,000 unique FoldingCorpus targets, with all 12 targets retained per protein and three
training seeds at every scale. Training each set for three epochs yields 21, 39, 96,
189, 375, 750, and 1,500 optimizer steps. The corresponding General-10 gains are
$0.49$, $0.61$, $1.37$, $2.92$, $3.31$, $3.70$, and $3.02\pp$
(Figure~\ref{fig:data_scaling}). Transfer therefore strengthens continuously from 50
to 2,000 proteins, peaks at $+3.70\pp$, and remains strong at $+3.02\pp$ with 4,000
proteins, indicating diminishing returns beyond 2,000 proteins under this schedule.

The source-side diagnostics follow a different profile. Held-out FoldingCorpus accuracy
rises from 0.478 at 50 proteins to 0.546 at 2,000 and becomes unstable at 4,000
(three-seed mean 0.418); FoldBench lDDT-C$\alpha$ increases from 0.244 to 0.252 and then
saturates, while Contact F1 keeps rising from 0.0666 to 0.0768. General-10 stays positive
despite the 4,000-protein instability, so downstream transfer is not determined by
endpoint FoldingCorpus accuracy alone.

Appendix~\ref{app:scaling_dynamics} reports all endpoint, seed-level, and checkpoint
results.

\subsection{Ablations: FoldingCorpus drives broad transfer while geometry concentrates on 3D}
\label{sec:ablations}

\begin{table}[H]
\centering
\caption{Component results on external benchmarks over three seeds. Values are mean
accuracies in percent; parentheses report absolute change from \base{} in percentage
points. The 3D macro averages FTB-Core, SpatialViz, and VSI; Text G7 averages the
remaining seven General-10 datasets.}
\label{tab:components}
\scriptsize
\setlength{\tabcolsep}{2.2pt}
\begin{tabular*}{\linewidth}{@{\extracolsep{\fill}}ccccccc@{}}
\toprule
\textbf{Arm} & \textbf{G10} & \textbf{3D macro} & \textbf{FTB} &
\textbf{SpatialViz} & \textbf{VSI} & \textbf{Text G7} \\
\midrule
\base{} & 45.09 & 40.44 & 36.18 & 26.61 & 58.53 & 47.09 \\
w/o FoldingCorpus & 45.77 (+0.68) & 41.47 (+1.03) & 38.44 (+2.26) &
27.40 (+0.79) & 58.58 (+0.05) & 47.61 (+0.52) \\
w/o Geometry & 48.02 (+2.93) & 43.30 (+2.86) & 39.00 (+2.81) &
31.81 (+5.20) & 59.08 (+0.54) & \textbf{50.05 (+2.96)} \\
\textbf{Fold2Reason-full} & \textbf{48.33 (+3.23)} & \textbf{44.38 (+3.94)} &
\textbf{40.26 (+4.08)} & \textbf{32.71 (+6.10)} & \textbf{60.16 (+1.63)} &
50.02 (+2.93) \\
\bottomrule
\end{tabular*}
\end{table}

\begin{figure}[H]
  \centering
  \includegraphics[width=0.93\linewidth]{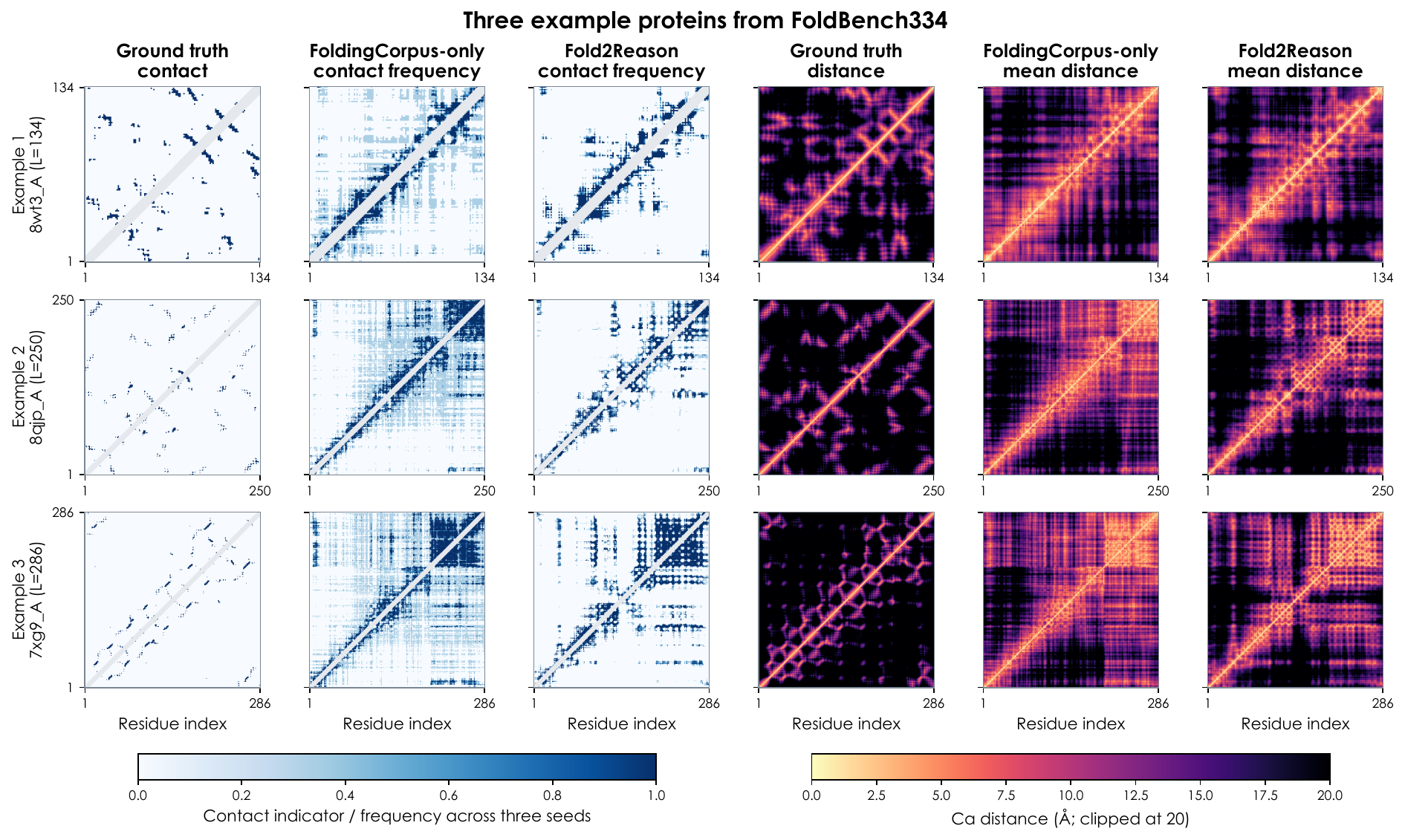}
  \caption{\textbf{Contact and distance maps for three example proteins.}
  Rows show three FoldBench334 proteins of increasing length (8wt3\_A, $L=134$;
  8qjp\_A, $L=250$; 7xg9\_A, $L=286$), comparing ground truth, FoldingCorpus-only, and
  Fold2Reason. Contacts use C$\alpha$ distance $<8$~\AA{} and sequence separation
  $\geq6$; predicted maps show frequency across all three seeds, with gray diagonal bands
  marking excluded near-sequence pairs. Distance maps show the three-seed mean, clipped
  at 20~\AA. Panels are unsmoothed and cover the complete protein.}
  \label{fig:structural_contact_distance_maps}
\end{figure}

The w/o Geometry arm raises General-10 by $2.93\pp$, so FoldingCorpus question
answering accounts for most of the external effect. Paired by seed, Fold2Reason-full
minus w/o Geometry is $0.57$, $0.49$, and $-0.15\pp$ (mean $+0.30\pp$); the sign
change across seeds makes this broad-transfer estimate seed-sensitive. That increment
is not spread evenly within General-10: relative to w/o Geometry, Fold2Reason-full
gains $1.26\pp$ on FTB-Core, $0.90\pp$ on SpatialViz, and $1.08\pp$ on VSI, raising
their 3D macro by $1.08\pp$ while changing Text G7 by $-0.03\pp$. This 3D-specific
difference is conditional on keeping the workspace; Pure LoRA has a slightly higher 3D
macro than Full (44.54\% versus 44.38\%).

The structural readout follows the complementary pattern
(Appendix~\ref{app:structural_audits}, Table~\ref{tab:structural_readout_complete}).
Relative to the workspace-equipped w/o Geometry arm, Fold2Reason-full increases mean
lDDT-C$\alpha$ by 0.00973 and Contact F1 by 0.00629, while TM-score decreases by
0.00214 and C$\alpha$ distance MAE increases by 0.148~\AA; the geometry-dominant w/o
FoldingCorpus arm preserves these readouts near the full model. The Geometry signal is
therefore concentrated in local structure, making residue neighborhoods and contacts
more recoverable from the shared representation while global topology does not improve
under this decoder. The absolute Full scores (TM-score 0.1688 and lDDT-C$\alpha$
0.2532) place this result in the regime of \emph{structural decodability}; competitive
protein folding requires substantially stronger global reconstruction. Read through its
native LM head on the same 334 targets, the unadapted Qwen3.5-9B backbone scores 4.55
TM-score, 9.38 lDDT-C$\alpha$, and 3.05 Contact F1 against 16.88, 25.32, and 6.90 for
\method{} (all scores $\times100$; Figure~\ref{fig:overview}a), and \method{} also leads
the best per-metric values of the four other general-purpose systems in that panel.
Appendix~\ref{app:geometry_distribution} reports the full-cohort distribution of the
paired per-protein changes and repeats the audit on three cases fixed at the lower,
median, and higher quantiles of the per-protein Contact F1 change before any map was
inspected.

Figure~\ref{fig:structural_contact_distance_maps} complements the aggregate distribution
with three individual proteins. Fold2Reason-full produces sharper and less diffuse
contact frequencies than FoldingCorpus-only in all three, and its mean distance maps show
more pronounced block structure, while both adapted models remain far from the
ground-truth global contact topology: Geometry sharpens selected local and pairwise
structure while global reconstruction remains unreliable.

Across the three arms, FoldingCorpus supervision governs broad behavioral transfer and
Geometry shapes local structural decodability and 3D-targeted behavior; the w/o
FoldingCorpus arm retains lDDT and contact readouts near Fold2Reason-full while
producing smaller FTB and SpatialViz gains.

\section{Discussion and Conclusion}

\method{} turns known protein structures into two training signals for a general
language model: verified FoldingCorpus answers through the native LM head and continuous
geometry through a frozen structural reader, with separable behavioral and
representational effects. FoldingCorpus answers account for most of the $3.23\pp$
General-10 gain, positive on every dataset mean. Geometry adds a smaller increment on
3D-oriented evaluations ($+1.26\pp$ on FTB-Core over the workspace-equipped w/o Geometry
arm) and local structural decodability, raising lDDT-C$\alpha$ and Contact F1 while
TM-score dips marginally and global folding quality stays low.

Matched controls locate the broader signal: hidden random-geometry targets improve the
aggregate, whereas format copying and fixed shuffled labels remain near zero on average.
Real protein structure is strongest among the tested sources, and the gap cannot be
reduced to learning the answer format alone. Across 50--4,000 proteins,
transfer peaks at 2,000 when data breadth and compute scale together, whereas raising labels
per protein from 3 to 12 at fixed coverage and optimizer steps gives no monotonic gain
(Appendix~\ref{app:label_bandwidth}): the evidence supports transfer from protein-derived
supervision, not label density. Whether this reflects reusable reasoning, more reliable
use of pretrained capabilities, or answer-format adaptation remains unresolved. Gains
extend to InternVL3.5-8B and Qwen3.5-2B, 4B, and 9B but not Gemma-4-12B-IT, implicating
architecture and adaptation dynamics.

These claims rest on 1,000 training proteins, rank-16 adapters, three seeds, and
General-10, and the frozen decoder measures structural decodability, not accurate
global reconstruction. Protein splits are cluster-disjoint, but FoldBench uses only
exact-sequence and 5-mer screening with known scaling-pool overlaps, leaving
remote-homology and novel-fold generalization unverified; task heterogeneity and
answer-format ambiguity further qualify behavioral gains
(Appendices~\ref{app:spatial_task_details}, \ref{app:evaluation_protocol},
and~\ref{app:scaling_dynamics}). Broader benchmarks and corpora, stronger
decoders, and more architectures come next. Within these limits, a solved scientific
problem becomes usable post-training data for general reasoning.

% ---------------------------------------------------------------------------
% ICLR 2027 required/encouraged statements. These are placed at the end of the
% main text, immediately before the bibliography, and do not count toward the
% page limit (see the ICLR 2027 Author Guidelines and AI Policy for Authors).
% ---------------------------------------------------------------------------

\bibliography{references}
\bibliographystyle{iclr2027_conference}

\appendix

\section{Training and Data Contract}
\label{app:training}

\begin{table}[h]
\centering
\caption{Canonical Fold2Reason-full training contract.}
\small
\begin{tabular*}{\linewidth}{@{\extracolsep{\fill}}cc@{}}
\toprule
\textbf{Item} & \textbf{Value} \\
\midrule
Base model & Qwen3.5-9B \\
Protein views & 336 sequence; 332 +MSA; 332 +MSA+template \\
Length range / mean & 29--199 / 144.42 residues \\
FoldingCorpus records & 12 per protein; 12,000 total \\
Packed supervision & 12 labels + EOS per protein \\
Workspace & width 256; 16 pooled tokens; at most 2,048 pairs \\
LoRA & rank 16; alpha 32; dropout 0.05 \\
Trainable parameters & 43,278,336 LoRA + 3,898,624 workspace \\
Frozen decoder & 3,308,332 parameters \\
Optimization & 3 epochs; 375 steps; 4 A800-80G GPUs \\
Seeds & 20260729, 20260803, 20260804 \\
Mean runtime & approximately 31.6 minutes per seed \\
\bottomrule
\end{tabular*}
\end{table}

\subsection{Optimization dynamics}
\label{app:optimization_dynamics}
Three instrumentation reruns use the canonical seeds, data order, rank-16 LoRA, shared workspace, and frozen decoder. Logged losses and gradient norms are training diagnostics, not checkpoint-selection criteria.

\subsection{Protein source, partitions, and structural preprocessing}

The source is the OpenFold monomer short-protein collection. We retain one selected,
relaxed monomer conformation per source ID and require standard amino acids, a
sequence--backbone length match, and finite N/CA/C/O atoms. The source parser accepts
only PDB \texttt{ATOM} records with blank or \texttt{A} alternate-location identifiers;
preprocessing excludes other chains, hetero atoms, side chains, and additional
conformations.
The target is centered and placed in a deterministic right-handed PCA-v1 frame, then
quantized at 0.1~\AA{} in the intermediate BB4Q10 record. Cache construction restores
floating-point N/CA/C/O coordinates. Pair distances and contacts are recomputed from
that cache and are invariant to translation and proper rotation; the right-handed
frame preserves chirality.

\begin{table}[h]
\centering
\caption{Protein data partitions and supervision counts. A FoldingCorpus record is one
operator applied to one protein; a packed sample contains all 12 records.}
\small
\begin{tabular*}{\linewidth}{@{\extracolsep{\fill}}ccccc@{}}
\toprule
\textbf{Partition} & \textbf{Proteins} & \textbf{Questions/protein} &
\textbf{FoldingCorpus records} & \textbf{Packed samples} \\
\midrule
Train & 1,000 & 12 & 12,000 & 1,000 \\
Development & 100 & 12 & 1,200 & 100 \\
Frozen test & 100 & 12 & 1,200 & 100 \\
\midrule
\textbf{Total} & \textbf{1,200} & \textbf{12} & \textbf{14,400} &
\textbf{1,200} \\
\bottomrule
\end{tabular*}
\end{table}

The 1,200 proteins occupy disjoint upstream MMseqs2 clusters \citep{steinegger2017mmseqs2} constructed at 30\%
sequence identity and 80\% coverage; source-ID and cluster overlap are zero for every
pair of partitions, and at most one protein is selected per cluster within a split.
These are checks of upstream cluster membership, not exhaustive pairwise or template isolation. Core-training versus FoldBench334 screens report zero exact-sequence matches and maximum 5-mer Jaccard 0.02381. Full MMseqs2, deposition-date, CATH, and SCOP isolation remain unverified \citep{sillitoe2019cath,fox2014scope}; the separate scaling-pool audit below reports nonzero template-source and alignment overlaps.

For train and development proteins, residues with pLDDT below 70 are excluded from
coordinate-derived supervision and the remaining residues receive confidence weights;
the median valid-residue fraction is 0.983. The frozen-test source uses its complete,
prefiltered backbone because its source artifact lacks the residue-level confidence
vector. Samples with fewer than $\max(3,\mathrm{round}(0.5L))$ valid residues are
rejected. No gap filling, chain stitching, alternate-conformer averaging, or side-chain
reconstruction is performed.

MSA views use the source OpenFold alignments (the retained artifacts include
\texttt{concat\_cfdb\_uniref100\_filtered.a3m} and \texttt{hmm\_output.sto}). The query
is row one; up to 31 additional unique rows are selected across sequence identity after
requiring at least 60\% coverage and identity in $[0.10,0.995]$. Template views use at
most four preassigned hits. We map finite template C$\alpha$ atoms to query indices,
retain pairs with sequence separation at least six and distance at most 12~\AA{}, and
encode the median distance in 0.5~\AA{} bins together with supporting-hit count, capped
at $4L$ pairs. Template identifiers and release dates are removed from model prompts.
No additional experiment-specific template-date cutoff was applied; the source
assignment retains release dates, so a temporal rerun must filter and regenerate the
template evidence before training.

\paragraph{Matched-control construction.}
\label{app:matched_control_details}
The three Table~\ref{tab:protein_specificity_controls} controls use Qwen3.5-9B
Pure-LoRA with the same 1,000 training packs, 12 answers per pack, rank-16 adapter,
three epochs (375 optimizer steps), and three seeds, without the workspace, Geometry
loss, or retrieval auxiliary. Fixed Shuffle preserves every protein input and question
but replaces each target with a fixed donor label drawn from a cluster-disjoint pack
with the same operator and candidate vocabulary; because label marginals are preserved
rather than forced to disagree, 42.85\% of training labels coincide with the native
target. Hidden Geometry instead derives all 12 targets in a pack from one unobserved
random point structure conditioned on the source length, residue mask, and radius of
gyration; the structure is sampled from an IID cloud, polymer chain, or clustered-shape
family, and only the 32-way retrieval question has its candidate fingerprints regenerated
so that one candidate represents the hidden target. Format Copy appends a donor
\texttt{Code=} label to each question and explicitly trains the model to copy these
operator-valid codes, making it a joint format, copying, and answer-prior control rather
than a strict format-only intervention. Control construction did not read General-10
items, answers, model errors, or scores. Thus, Hidden Geometry tests internally coherent
but input-independent geometric targets, Fixed Shuffle tests fixed mismatched
supervision, and Format Copy estimates the transfer available from producing valid short
answers without solving the structural questions.

\section{FoldingCorpus Construction}
\label{app:foldingcorpus_operators}

Let $c_i\in\mathbb{R}^3$ be the valid C$\alpha$ coordinate of residue $i$,
$d(i,j)=\lVert c_i-c_j\rVert_2$, and
$\bar c=|V|^{-1}\sum_{i\in V}c_i$. Every protein contributes exactly one record for
each operator in Table~\ref{tab:foldingcorpus_operators}. A salted deterministic hash first
chooses the intended class and candidate order; the generator then samples a valid
coordinate tuple for that class. When a requested class has no valid tuple, it falls
back to the available candidate set and records the resulting label. Distance-order
records sample from the lowest and highest distance deciles before independently
permuting option order. These rules prevent answer position from being a deterministic
function of the operator while preserving one record per operator and protein.

\begin{table}[h]
\centering
\caption{The 12 FoldingCorpus operators. Counts are observed labels in the 1,000-protein
training partition. Indices refer to valid residues; \texttt{sep} is sequence
separation.}
\label{tab:foldingcorpus_operators}
\scriptsize
\setlength{\tabcolsep}{2.2pt}
\begin{tabularx}{\linewidth}{>{\centering\arraybackslash}p{0.18\linewidth}>{\centering\arraybackslash}X>{\centering\arraybackslash}p{0.23\linewidth}}
\toprule
\textbf{Operator / train labels} & \textbf{Definition and threshold} &
\textbf{Sampling rule} \\
\midrule
CONTACT\_\allowbreak SHORT A470/B530 & $A$ iff $d(i,j)<8$~\AA{} & $6\leq\mathrm{sep}\leq11$ \\
CONTACT\_\allowbreak MEDIUM A458/B542 & $A$ iff $d(i,j)<8$~\AA{} & $12\leq\mathrm{sep}\leq23$ \\
CONTACT\_\allowbreak LONG A447/B553 & $A$ iff $d(i,j)<8$~\AA{} & $\mathrm{sep}\geq24$ \\
DISTANCE\_\allowbreak ORDER\_\allowbreak 1 A481/B519 & $A$ iff $d(i,j)<d(k,l)$ & low/high distance deciles; local/medium pairs \\
DISTANCE\_\allowbreak ORDER\_\allowbreak 2 A512/B488 & $A$ iff $d(i,j)<d(k,l)$ & low/high distance deciles; long-range pairs \\
SEGMENT\_\allowbreak ORIENTATION A331/B297/C372 & For endpoint vectors $u,v$: A if $\cos(u,v)\geq0.5$; B if $\cos(u,v)\leq-0.5$; C otherwise & segment starts separated by at least eight residues \\
NEARER\_\allowbreak CENTER A491/B509 & A iff $\lVert c_i-\bar c\rVert_2<\lVert c_j-\bar c\rVert_2$ & two valid residues; option order permuted \\
LOCAL\_\allowbreak FRAME\_\allowbreak DIRECTION A508/B492 & A iff $(c_j-c_i)^\top(C_i-CA_i)\geq0$ & source/target separation at least six \\
CA\_\allowbreak CHIRALITY A548/B452 & A iff $[(c_2-c_1)\times(c_3-c_1)]^\top(c_4-c_1)\geq0$ & four consecutive valid C$\alpha$ atoms \\
MULTI\_\allowbreak CONSTRAINT A304/B317/C379 & A/B selects the nearer candidate only if its distance is $<8$~\AA{}; C if neither qualifies & one anchor and two valid candidates \\
RETRIEVAL\_\allowbreak 32 21--38/label & Select the target fingerprint among labels A--Z,a--f & target plus 31 same-split hard negatives; order permuted \\
CONTACT\_\allowbreak ANY A454/B546 & A iff $d(i,j)<8$~\AA{} & any pair with $\mathrm{sep}\geq6$ \\
\bottomrule
\end{tabularx}
\end{table}

The 32-way fingerprint contains sequence length, radius of gyration, dominant
secondary-structure class, and eight representative long-range contacts quantized to
a $32\times32$ index grid. Hard negatives are nearest neighbors under length, radius,
secondary-structure fractions, contact density and span, mean contact degree, and two
covariance-shape ratios. All 32 labels occur in train (21--38 times each).

Nine operators use labels A/B, SEGMENT\_ORIENTATION and MULTI\_CONSTRAINT use A/B/C,
and RETRIEVAL\_32 uses A--Z,a--f. Tokenizer audit confirms that every label is one
token. Thus each packed sample supervises 12 answer tokens and EOS. The generator
independently recomputes every answer from coordinates; all 14,400 exported records
pass schema, candidate, split, and answer-recomputation checks.

\section{FTB-Core Construction and Audit}
\label{app:ftb_core}

FTB-Core v1.0.0 is generated deterministically by
\texttt{scripts/generate\_ftb\_core.py} (generator SHA-256 prefix
\texttt{52b2b1cf}) and was frozen on 2026-07-24. It contains 36,000 train, 6,000
validation, 6,000 test, and 6,000 size/range-shifted test-OOD examples. The paper uses
all 12,000 test and test-OOD examples, with 500 examples from each split for each of
the 12 tasks.

\begin{table}[h]
\centering
\caption{FTB-Core v1 task composition. The final column gives one prompt schema per
task; each task contributes 500 test and 500 test-OOD examples to the headline score.}
\scriptsize
\begin{tabularx}{\linewidth}{>{\centering\arraybackslash}p{0.25\linewidth}>{\centering\arraybackslash}p{0.18\linewidth}>{\centering\arraybackslash}X}
\toprule
\textbf{Task} & \textbf{Answer/parser} & \textbf{Representative generated query} \\
\midrule
nearest\_\allowbreak neighbor & point ID / exact & choose the nearest labeled 3D point \\
bond\_\allowbreak angle & degrees / $\pm0.5$ & compute an angle from three coordinates \\
signed\_\allowbreak dihedral & degrees / $\pm0.5$ & compute a signed four-point dihedral \\
tetrahedral\_\allowbreak chirality & positive/negative & determine the handedness of four points \\
proper\_\allowbreak rigid\_\allowbreak equivalence & yes/no & distinguish a proper rigid transform from reflection \\
sparse\_\allowbreak constraints\_\allowbreak candidate\_\allowbreak selection & A--D & choose the structure satisfying sparse distances \\
distance\_\allowbreak constraint\_\allowbreak audit & constraint ID/none & identify the violated distance constraint \\
sphere\_\allowbreak collision & yes/no & test collision between translated spheres \\
straight\_\allowbreak path\_\allowbreak clearance & yes/no & test whether a swept path clears obstacles \\
noisy\_\allowbreak template\_\allowbreak selection & A--D & select a candidate from noisy distance evidence \\
weighted\_\allowbreak contact\_\allowbreak evidence & contact/no\_contact & aggregate confidence-weighted contact evidence \\
fragment\_\allowbreak interface\_\allowbreak assembly & A--D & select the assembly satisfying interface constraints \\
\bottomrule
\end{tabularx}
\end{table}

Generation seeds are 101/202/303/404 for train/validation/test/test-OOD. The manifest
stores split hashes, stable IDs, task counts, answer types, and tolerances. Validation
recomputes file hashes, required fields, unique IDs across splits, finite numeric
targets, answer-schema validity, and exact per-task counts over all 54,000 records.
An additional SHA-256 audit over the rendered prompt strings finds no duplicate within
any split and zero exact-prompt overlap for every pair of splits. This audit eliminates
exact prompt reuse; semantic near-duplicates remain possible under the shared
generator. Answer correctness is exhaustively program-audited because labels are
deterministic outputs. Prompt clarity lacks separately logged human validation, which
limits the benchmark audit.
The release will include the generator, all rendered prompts, one complete rendered
example for every task, split manifests, parser, and validation script. All model
training excludes every FTB-Core split.

\section{General-10 Evaluation Protocol}
\label{app:evaluation_protocol}

Table~\ref{tab:evaluation_contract} fixes the local dataset revisions and headline
sample counts. Full 40-character revisions and file hashes are retained in each
\texttt{SOURCE.json} and the sealed selection manifests.

\begin{table}[h]
\centering
\caption{Per-benchmark evaluation contract. All rows use greedy decoding. EM denotes
the benchmark-specific normalized exact matcher; MCA is mean relative accuracy.}
\label{tab:evaluation_contract}
\scriptsize
\setlength{\tabcolsep}{2.0pt}
\begin{tabular*}{\linewidth}{@{\extracolsep{\fill}}ccccc@{}}
\toprule
\textbf{Benchmark (revision)} & \textbf{$n$} & \textbf{Modality} &
\textbf{New tokens} & \textbf{Score / extraction} \\
\midrule
FTB-Core v1.0.0 & 12,000 & text & 32 & numeric tolerance or categorical EM \\
SpatialViz (\texttt{f38482a8}) & 1,180 & image+text & 128 & option-letter accuracy \\
VSI (\texttt{d7cb1a39}) & 5,130 & video+text & 16 & MC accuracy/MCA; aggregation below \\
GraphQA Easy (\texttt{18f55c29}) & 21,600 & text & 128 & canonical answer-tail EM \\
GraphQA Hard (\texttt{b7e910dd}) & 21,600 & text & 128 & canonical answer-tail EM \\
BBH (\texttt{982bb89f}) & 6,511 & text & 128 & normalized EM \\
ChemBench (\texttt{6e1d2574}) & 2,785 & text & 128 & deterministic option-letter accuracy \\
ChemBench4K (\texttt{9b8fec19}) & 4,009 & text & 128 & option-letter accuracy \\
Lab-Bench (\texttt{5c77cec6}) & 1,967 & text & 128 & option-letter accuracy \\
SciBench (\texttt{93931252}) & 692 & text & 128 & last number, 1\% relative tolerance \\
\bottomrule
\end{tabular*}
\end{table}

VSI question scores use the official multiple-choice and numeric scoring rules.
The canonical endpoint and component results average these scores over all 5,130
questions. The independent scaling study retains its archived eight-task macro;
Appendix~\ref{app:scaling_numeric_details} gives both evaluation contracts and their
frozen base scores. GraphQA in the canonical Qwen9 results uses the corrected
answer-tail matcher. The model-family tables retain their archived within-model
scoring contracts, detailed in Appendix~\ref{app:model_family_details}.

\paragraph{Exact prompt construction.}
The sealed Qwen text runs use the system message ``Answer the benchmark item
accurately. Return only the final answer, without explanation.'' The FTB-Core system
message is ``Solve the spatial reasoning problem carefully. Return only the final
answer requested by the problem, without explanation.'' BBH uses the benchmark
\texttt{input} verbatim, and GraphQA uses \texttt{question} verbatim. ChemBench,
ChemBench4K, and Lab-Bench items with distractors use
\texttt{question}, a blank line, ``Options:'', one \texttt{LETTER. choice} per line,
and ``Return only the option letter.'' Lab-Bench items without distractors use the
question verbatim. SciBench uses \texttt{problem\_text}. These strings are passed
through the model's official chat
template with thinking disabled and an assistant-generation marker.

SpatialViz supplies the original image and renders the question and four choices after
the instruction to return one letter inside \texttt{<answer></answer>}. VSI prefixes
``These are frames of a video.''; multiple-choice items append the listed options and
request the option letter, while numeric items request one word or phrase. VSI resolves
288 scene videos and uniformly samples 32 frames per video; all questions for a scene
reuse the same frame tensor. Qwen uses its official image/video processor.

\paragraph{Model-family differences.}
InternVL and Gemma use their tokenizer-native chat templates and official media
processors. Selected IDs, benchmark user content, generation caps, and scorers remain
fixed within each base--adapter comparison. Gemma text evaluation uses a semantically equivalent deterministic answer-only
system instruction. Its SpatialViz and VSI wrappers seed the assistant response with
\texttt{<answer>} and \texttt{Final answer:}, respectively, because the native chat
interface otherwise continued with explanatory text. These fixed openings contain no correct option or target value and apply to both Base and adapters. The same parser strips them, but no with/without-prefix ablation is available. Appendix~\ref{app:output_validity} reports parse-failure rates, valid-output-only accuracy, and common-valid paired comparisons for Qwen3.5-9B Base and the three canonical Full adapters on FTB-Core and SpatialViz. Equivalent audits are not available for the other model-family settings and benchmarks; exact truncation rates cannot be recovered from decoded-text-only archives and must not be interpreted as zero. These diagnostics distinguish failed extraction from valid but incorrect answers, but do not fully separate reasoning improvements from answer-format or answer-selection changes.

Decoding is deterministic (\texttt{do\_sample=False}, no beams, no benchmark-specific
few-shot demonstrations). Parsers first discard generated role continuations. A
missing option, malformed categorical label, absent numeric value, non-finite number,
or output outside the task tolerance receives zero credit. GraphQA additionally
accepts an exact normalized final-answer tail, matching numeric suffix, or final
yes/no/true/false/unknown label; the same canonical matcher is applied to Base and all
adapters. SpatialViz uses strict option extraction. VSI uses accuracy for multiple
choice and the official 0.50--0.95 mean-relative-accuracy thresholds for numeric
questions. Canonical endpoint results average all questions; the independent scaling study uses the archived eight-task macro, as specified above.

\section{Geometry Decoder Pretraining Protocol}
\label{app:geometry_decoder}

The geometry decoder is pretrained in a Phase-0 head-only stage and reused as a fixed
readout in all component arms. In Phase 0, Qwen3.5-9B and all LoRA parameters were
frozen; the coordinate head
(2,112,012 parameters) and distogram head (1,196,320 parameters) were optimized, for
3,308,332 trainable decoder parameters in total. The decoder was trained for three
epochs, 375 optimizer steps, on the OpenFold high-confidence 1K training split
(336 sequence-only, 332 sequence+MSA, and 332 sequence+MSA+template proteins), using
the same coordinate, pair-distance, contact, distogram, local-frame, torsion, and
radius-of-gyration losses used by the geometry objective.

\begin{table}[h]
\centering
\caption{Frozen geometry decoder provenance.}
\small
\setlength{\tabcolsep}{4pt}
\begin{tabularx}{\linewidth}{>{\centering\arraybackslash}p{0.28\linewidth}>{\centering\arraybackslash}X}
\toprule
\textbf{Item} & \textbf{Protocol} \\
\midrule
Initialization & Pretrained Phase-0 head-only decoder \\
Trainable in Phase 0 & Coordinate and distogram heads only; base LM and LoRA frozen \\
Phase-0 data & 1,000 training proteins from the OpenFold high-confidence split \\
Fold2Reason training & Decoder loaded with \texttt{requires\_grad=False} and excluded from optimization \\
Held-out exclusions & No FoldBench334, General-10, FTB-Core, SpatialViz, VSI, GraphQA, BBH,
ChemBench, ChemBench4K, Lab-Bench, or SciBench examples \\
Extra structural data & No additional PDB-derived proteins beyond the 1K training split \\
External evaluation & Decoder absent; downstream benchmarks use only the adapted Qwen LM \\
\bottomrule
\end{tabularx}
\end{table}

This design fixes the geometry readout across arms. During \method{} training,
$g_{\psi}$ receives the residue workspace output $E$ as its sole input. Target
coordinates and benchmark inputs remain outside the decoder path. Trainability checks,
optimizer exclusion, initial/final parameter hashes, and gradient/update audits verify
that decoder parameters remain fixed. The decoder supplies a structural training loss
and is removed for external evaluation.

The Phase-0 decoder carries a structural readout prior learned from the 1K training
proteins and Qwen's existing residue-marker representations. FoldBench and geometry
losses therefore measure structural decodability and support the mechanism analysis.
General-10 measures behavioral transfer from the adapted language model after the
protein and decoder interfaces are removed.

\begingroup
\raggedbottom
\section{Algorithmic Specification and Worked Data Examples}
\label{app:algorithms_examples}

This section specifies the data-to-update path and illustrates it with an actual
training protein. Algorithms~\ref{alg:foldingcorpus_cache}--\ref{alg:readout_protocol}
cover target construction, post-training, and evaluation. The worked example in
Appendix~\ref{app:worked_protein} links a source record to the packed answer tokens
and loss mask used by the implementation.

\floatstyle{ruled}
\newfloat{f2rAlgorithm}{tbp}{loa}
\floatname{f2rAlgorithm}{Algorithm}

\subsection{Constructing the frozen training cache}

Algorithm~\ref{alg:foldingcorpus_cache} operates on the training partition after the
structural preprocessing in Appendix~\ref{app:training}. Coordinates determine the
answers offline. The cached protein prompt contains sequence and its assigned
optional evidence view; the FoldingCorpus prompt contains sequence, questions, and
candidate descriptions. Numerical answer evidence and source-ID mappings remain
in the audit record. Independent hash salts control operator labels, option order,
and packed question order; the completed packs are reused across training epochs.

\begin{f2rAlgorithm}[H]
\caption{Construct a FoldingCorpus training cache}
\label{alg:foldingcorpus_cache}
\small
\begin{tabularx}{\linewidth}{@{}rX@{}}
 & \textbf{Input:} training proteins $(s,v,Y,m,w)$, tokenizer $T$, fixed data seed $s_d$; $v$ is the assigned evidence view. \\
 & \textbf{Output:} frozen protein inputs $x$, marker indices $I$, FoldingCorpus tokens $z$, masked labels $\ell$, and geometry targets. \\
1 & Compute valid C$\alpha$ distances and structural summaries from $Y,m$. \\
2 & Build the training-pool hard-negative index used by the 32-way summary question. \\
3 & \textbf{For each} protein $p$ in the training partition: \\
4 & \quad Serialize $(s_p,v_p)$ and a residue skeleton with exactly one marker per residue; tokenize to obtain $(x_p,I_p)$. \\
5 & \quad \textbf{For each} of the 12 operators in Table~\ref{tab:foldingcorpus_operators}: \\
6 & \qquad Choose its intended class with an operator-specific salted hash; use the protein-seeded RNG for argument and option sampling. \\
7 & \qquad Sample valid arguments under the operator's separation and class rules; use the available-candidate fallback when needed. \\
8 & \qquad Construct question $q_k$, candidate labels, and answer $a_k$ from $Y_p,m_p$; retain numerical evidence separately. \\
9 & \quad Independently recompute every answer and check candidate membership. \\
10 & \quad Permute the 12 question--answer pairs using the protein-specific training-order salt. \\
11 & \quad Render the sequence and numbered questions with the chat template; append the space-separated answers and EOS to obtain $z_p$. \\
12 & \quad Set $\ell_p=-100$ on every prompt position and $\ell_p=z_p$ on the answer/EOS positions. \\
13 & \quad Check one marker per residue and 13 supervised tokens; save $(x_p,I_p,z_p,\ell_p,Y_p,m_p,w_p)$ and audit metadata. \\
\end{tabularx}
\end{f2rAlgorithm}

The 32-way operator presents one target summary and 31 hard-negative summaries in
permuted order. Its one-token answer is part of the same FoldingCorpus CE as the other
11 answers. The canonical Full recipe disables the separate fingerprint-projection
retrieval objective; that setting does not remove the textual 32-way question.

\subsection{Post-training with shared and language-only paths}

Let $\theta_0$ denote the frozen base weights, $\Delta\theta$ the LoRA parameters,
$\phi$ the active workspace parameters, and $\psi$ the pretrained geometry decoder.
Algorithm~\ref{alg:full_training} describes the canonical Full run. Both forwards
use the same adapted language model. The workspace and frozen decoder remain in
the gradient path, while the optimizer updates only $\Delta\theta$ and $\phi$.

\begin{f2rAlgorithm}[H]
\caption{Canonical Full post-training and its Pure-LoRA specialization}
\label{alg:full_training}
\small
\begin{tabularx}{\linewidth}{@{}rX@{}}
 & \textbf{Input:} frozen cache, base $\theta_0$, pretrained decoder $g_\psi$, training seed, three-epoch schedule. \\
 & \textbf{Output:} LoRA adapter $\Delta\theta$ and workspace $\phi$; unchanged decoder $\psi$. \\
1 & Initialize rank-16 LoRA and workspace; freeze $\theta_0$, $\psi$, and the unused retrieval projection. \\
2 & Create AdamW groups for LoRA and active workspace parameters only. \\
3 & \textbf{For each} epoch and seed-shuffled gradient-accumulation window: \\
4 & \quad Zero optimizer gradients; distribute protein packs across workers. \\
5 & \quad \textbf{For each} active protein pack $(x,I,z,\ell,Y,m,w)$: \\
6 & \qquad $H\leftarrow f_{\theta_0,\Delta\theta}(x)_I$; retain the computation graph. \\
7 & \qquad $(E,M)\leftarrow W_\phi(H)$, with residue-aligned $E$ and 16 prefix embeddings $M$. \\
8 & \qquad $(\widehat Y,\widehat D)\leftarrow g_\psi(E)$; $\widehat D$ contains pairwise distogram logits. \\
9 & \qquad $\mathcal L_{\mathrm{geo}}\leftarrow\mathrm{GeometryLoss}(\widehat Y,\widehat D;Y,m,w)$, using the seven terms in Section~\ref{sec:method}. \\
10 & \qquad $u\leftarrow[M;\mathrm{Embed}_{\theta_0}(z)]$ and $\widetilde\ell\leftarrow[-100^{16};\ell]$. \\
11 & \qquad $\mathcal L_{\mathrm{qa}}\leftarrow\mathrm{CausalCE}(f_{\theta_0,\Delta\theta}(u),\widetilde\ell)$; average over the 13 supervised positions. \\
12 & \qquad Backpropagate the accumulation-window-normalized $\mathcal L_{\mathrm{qa}}+\mathcal L_{\mathrm{geo}}$. \\
13 & \quad Synchronize gradients, clip their norm to 1.0, update LoRA/workspace with AdamW, and advance the learning-rate scheduler. \\
14 & Save declared checkpoints and the fixed three-epoch endpoint; verify the decoder is unchanged. \\
15 & \textbf{Pure-LoRA specialization:} initialize only LoRA, omit workspace/decoder and lines 6--9, set $u=\mathrm{Embed}_{\theta_0}(z)$ and $\widetilde\ell=\ell$, and optimize only $\Delta\theta$ with $\mathcal L_{\mathrm{qa}}$. \\
\end{tabularx}
\end{f2rAlgorithm}

The workspace first maps $H$ to width 256. It forms pairs at sequence offsets
1--4 and fills the remaining budget of at most 2,048 pairs with evenly spaced
indices from longer-range pairs. Symmetric pair features concatenate absolute differences and
elementwise products; messages are averaged at their incident residues. Residual
transitions produce the updated residue features. A residual projection returns
$E$ at LM width, while 16 learned queries attention-pool the reduced features and
project the pooled vectors to $M$. Pair construction depends on sequence indices,
not target contacts.

For the canonical 1,000-protein run, four workers each process one protein per
microstep and accumulate two microsteps, giving eight proteins per optimizer step
and 375 steps over three epochs. LoRA and workspace learning rates are $10^{-4}$
and $3\times10^{-4}$, with weight decay 0.01, a 5\% warmup, and cosine decay.
LoRA uses alpha 32 and dropout 0.05. These values describe the canonical run;
the scaling runs use their declared data sizes and exposure schedules.

In implementation terms, freezing the decoder sets its parameters to
\texttt{requires\_grad=False} and excludes them from optimizer groups. Its forward
pass is still differentiable with respect to $E$. This distinction preserves
geometry gradients into the workspace and LoRA. The Pure-LoRA specialization uses
the same packed FoldingCorpus tensors, without a workspace or geometry forward pass.

\subsection{Packed data and supervision masks}
\label{app:worked_protein}\label{app:worked_records}\label{app:worked_mask}
Each protein contributes one packed prompt with 12 structural answers and EOS. Prompt tokens are masked from the answer loss. Full prepends 16 workspace embeddings; Pure LoRA uses the same cached answer sequence without those embeddings or the protein-stream forward pass. The operator definitions and Algorithms~\ref{alg:foldingcorpus_cache} and~\ref{alg:full_training} specify target construction and training. 
\subsection{Evaluation interfaces and aggregation}

The three evaluation interfaces answer different questions. FoldingCorpus evaluation
tests the learned one-token decisions, FoldBench evaluates the coordinate readout,
and General-10 evaluates the adapted language model on downstream inputs.
Algorithm~\ref{alg:readout_protocol} makes the inputs and aggregation explicit.

\begin{f2rAlgorithm}[H]
\caption{Evaluate fixed checkpoints on FoldingCorpus, FoldBench, and General-10}
\label{alg:readout_protocol}
\small
\begin{tabularx}{\linewidth}{@{}rX@{}}
 & \textbf{Input:} declared checkpoint(s), frozen evaluation manifests, prompts, decoding settings, and scorers. \\
 & \textbf{Output:} per-example predictions, per-seed metrics, and seed-aggregated results. \\
1 & \textbf{For each} training seed, load its declared checkpoint in evaluation mode. \\
2 & \textbf{Held-out FoldingCorpus questions:} for each protein, compute its matched workspace memory for Full; omit memory for Pure-LoRA. \\
3 & \quad Ask each question independently using the sequence and that question; do not append earlier gold answers. \\
4 & \quad Take the first next-token argmax over the complete vocabulary and compare with the canonical label token. \\
5 & \quad Average correctness within each operator, then average the 12 operator accuracies. Record candidate-restricted accuracy separately as a diagnostic. \\
6 & \textbf{FoldBench:} for each of the 334 proteins, run the protein stream, workspace, and fixed coordinate decoder for the structural-readout arm. \\
7 & \quad Score its predicted coordinates against the valid reference residues; average each metric over the 334 proteins within the seed. \\
8 & \textbf{General-10:} load only the base LM plus LoRA adapter; remove workspace, protein inputs, and geometry decoder. \\
9 & \quad Evaluate every example in each of the ten fixed manifests with the dataset's input modality, decoding contract, and scorer. \\
10 & \quad Pair adapted and base results by sample ID under identical prompts and scoring; compute each dataset score $a_{s,d}$ and difference $\delta_{s,d}$. \\
11 & \quad Compute the per-seed macro gain $\Delta_s=10^{-1}\sum_{d=1}^{10}\delta_{s,d}$. \\
12 & Aggregate per-seed scores and gains with the reported mean and seed variability; retain every seed and the declared checkpoints. \\
\end{tabularx}
\end{f2rAlgorithm}

For FoldingCorpus accuracy, an out-of-set first token is incorrect even if the correct
label has the highest score among the listed options. For structural metrics,
each seed's coordinates are scored before averaging scores across seeds;
averaging distance/contact maps is a visualization operation. General-10 scores
are expressed as percentages, so $\delta_{s,d}$ and $\Delta_s$ are percentage-point
changes. Each dataset has equal weight, independent of its number of examples. The detailed
benchmark parsing and decoding contract remains in Appendix~\ref{app:evaluation_protocol}.

\endgroup

\section{Model-Scale and Model-Family Results}
\label{app:model_family_details}

The model-family comparison evaluates the transfer of FoldingCorpus supervision through
Pure LoRA. All five settings train on 1,000 proteins and the same twelve-question FoldingCorpus
inventory. All models use three epochs and 375 optimizer steps. Each comparison pairs an adapted checkpoint with its own base model and its archived prompt and scoring contract.

The saved configurations verify rank 16, alpha 32, dropout 0.05, and zero trainable
parameters outside LoRA for all fifteen runs. The Qwen adapters cover the language
decoder's attention, linear-attention projections, and MLP projections. InternVL
and Gemma use the language-decoder q/k/v/o and gate/up/down projections available
in their respective architectures. Vision modules and multimodal projectors remain
frozen. The source package records the exact module counts for each seed.

\begin{table}[H]
\centering
\caption{Model-family training contracts from the saved run configurations. All arms use rank-16, alpha-32 LoRA with dropout 0.05; the parameter count includes only trainable adapters.}
\label{tab:ae_model_contract}
\scriptsize
\setlength{\tabcolsep}{3pt}
\begin{tabular*}{\linewidth}{@{\extracolsep{\fill}}lrrrrr@{}}
\toprule
Model & Epochs & Steps & Peak LR & LoRA (M) & GPUs/seed \\
\midrule
Qwen3.5-2B & 3 & 375 & 1e-04 & 16.82 & 4 \\
Qwen3.5-4B & 3 & 375 & 1e-04 & 32.46 & 4 \\
Qwen3.5-9B & 3 & 375 & 1e-04 & 43.28 & 4 \\
InternVL3.5-8B & 3 & 375 & 1e-04 & 43.65 & 4 \\
Gemma-4-12B-IT & 3 & 375 & 1e-04 & 65.57 & 4 \\
\bottomrule
\end{tabular*}
\vspace{3pt}
\parbox{0.98\linewidth}{\footnotesize All three configurations per model were checked. Each model uses 1,000 training proteins, all 12 FoldingCorpus labels, and zero non-LoRA trainable parameters.}
\end{table}

All three Qwen sizes and InternVL improve in each of the three seeds. Their
mean General-10 gains are $5.414$, $4.844$, $3.223$, and $1.529\pp$.
Gemma's changes are $-0.690$, $+0.509$, and $+0.398\pp$, giving a mean of
$+0.072\pp$ and a t interval spanning zero. These seed distributions support
the positive transfer and family dependence reported in the main text.

\begin{table}[H]
\centering
\caption{General-10 gains by model and training seed. Each row is paired with its own base checkpoint and archived evaluation contract.}
\label{tab:ae_model_uncertainty}
\scriptsize
\setlength{\tabcolsep}{3pt}
\begin{tabular*}{\linewidth}{@{\extracolsep{\fill}}lrrrrr@{}}
\toprule
Model & S1 & S2 & S3 & Mean $\pm$ SD & 95\% t interval \\
\midrule
Qwen3.5-2B & +6.448 & +5.048 & +4.747 & $5.414 \pm 0.907$ & [+3.160, +7.668] \\
Qwen3.5-4B & +4.636 & +5.155 & +4.742 & $4.844 \pm 0.274$ & [+4.164, +5.525] \\
Qwen3.5-9B & +3.216 & +3.204 & +3.249 & $3.223 \pm 0.024$ & [+3.164, +3.281] \\
InternVL3.5-8B & +1.754 & +1.225 & +1.607 & $1.529 \pm 0.273$ & [+0.851, +2.206] \\
Gemma-4-12B-IT & -0.690 & +0.509 & +0.398 & $0.072 \pm 0.663$ & [-1.574, +1.718] \\
\bottomrule
\end{tabular*}
\vspace{3pt}

\end{table}

\subsection{Complete dataset profiles}
The table reports every benchmark and every model, including negative means. The nine-billion-parameter Qwen reference uses Pure LoRA and the corrected answer-tail GraphQA scorer. Other model rows retain their archived within-model scoring contracts. This is a set of within-model comparisons, not a common-budget ranking.

% BEGIN INLINED: revision-evidence/model_profiles.tex
\begin{table}[htbp]
\centering
\caption{Complete Pure-LoRA mean changes from each model's own base (percentage points). All benchmarks and model settings are retained.}
\label{tab:compact_models}
\scriptsize
\setlength{\tabcolsep}{3pt}
\begin{tabular*}{\linewidth}{@{\extracolsep{\fill}}lrrrrr@{}}
\toprule
Benchmark & Qwen 2B & Qwen 4B & Qwen 9B & InternVL 8B & Gemma 12B \\
\midrule
FTB-Core & +10.45 & +7.46 & +4.10 & +2.64 & -1.23 \\
SpatialViz & +21.21 & +22.20 & +6.69 & -0.45 & -0.79 \\
VSI & -0.28 & +3.99 & +1.50 & +0.20 & +2.95 \\
GraphQA Easy & +2.80 & -2.82 & +1.96 & +0.22 & -0.43 \\
GraphQA Hard & +6.51 & +3.69 & +9.95 & +3.42 & -1.93 \\
BBH & +1.35 & +2.91 & +2.38 & +1.35 & +1.53 \\
ChemBench & +0.08 & +3.26 & +0.33 & +3.29 & +0.87 \\
ChemBench4K & +6.12 & +2.80 & +2.99 & +0.35 & -1.10 \\
Lab-Bench & +2.86 & +2.19 & +1.02 & +0.31 & +0.97 \\
SciBench & +3.05 & +2.76 & +1.32 & +3.97 & -0.11 \\
General-10 & +5.41 & +4.84 & +3.22 & +1.53 & +0.07 \\
\bottomrule
\end{tabular*}
\vspace{3pt}
\end{table}
% END INLINED: revision-evidence/model_profiles.tex

\section{Task-Level Decomposition of Spatial Transfer}
\label{app:spatial_task_details}
The compact table retains all 12 FTB-Core and all 12 SpatialViz tasks, including zero and negative changes. Full improves FTB-Core test and test-OOD by $3.94$ and $4.21\pp$; the largest family gain is constraint satisfaction ($12.19\pp$). SpatialViz category means are positive, but constituent tasks differ markedly.
% BEGIN INLINED: revision-evidence/spatial_tasks.tex
\begin{table}[htbp]
\centering
\caption{Complete task-level mean scores for Full. Scores are percentages; changes are percentage points. Abbreviated FTB task names follow the construction table.}
\label{tab:compact_tasks}
\scriptsize
\setlength{\tabcolsep}{3pt}
\begin{tabular*}{\linewidth}{@{\extracolsep{\fill}}lrrrlrrr@{}}
\toprule
FTB-Core task & Base & Full & $\Delta$ & SpatialViz task & Base & Full & $\Delta$ \\
\midrule
Bond angle & 0.60 & 0.23 & -0.37 & 2DRotation & 0.00 & 5.00 & +5.00 \\
Distance audit & 9.40 & 7.50 & -1.90 & 3DRotation & 30.00 & 35.42 & +5.42 \\
Fragment assembly & 28.30 & 30.10 & +1.80 & 3ViewProjection & 31.00 & 38.67 & +7.67 \\
Nearest neighbor & 41.50 & 50.47 & +8.97 & ArrowMoving & 30.00 & 33.75 & +3.75 \\
Noisy template & 25.80 & 26.97 & +1.17 & BlockMoving & 31.25 & 31.25 & +0.00 \\
Rigid equivalence & 47.20 & 48.20 & +1.00 & CrossSection & 14.17 & 11.39 & -2.78 \\
Signed dihedral & 0.40 & 0.37 & -0.03 & CubeAssembly & 1.25 & 50.00 & +48.75 \\
Sparse selection & 32.50 & 69.17 & +36.67 & CubeCounting & 40.00 & 44.72 & +4.72 \\
Sphere collision & 53.60 & 55.80 & +2.20 & CubeReconstruction & 30.00 & 30.56 & +0.56 \\
Path clearance & 49.30 & 48.70 & -0.60 & CubeUnfolding & 33.33 & 30.00 & -3.33 \\
Chirality & 49.40 & 49.37 & -0.03 & MechanicalSystem & 43.75 & 52.92 & +9.17 \\
Contact evidence & 96.20 & 96.23 & +0.03 & PaperFolding & 27.50 & 33.89 & +6.39 \\
\bottomrule
\end{tabular*}
\vspace{3pt}
\parbox{0.98\linewidth}{\footnotesize Each FTB task has 1,000 questions. SpatialViz counts are 80 each for 2DRotation, 3DRotation, ArrowMoving, BlockMoving, CubeAssembly, and MechanicalSystem; 100 for 3ViewProjection; and 120 each for the other five tasks.}
\end{table}
% END INLINED: revision-evidence/spatial_tasks.tex

\paragraph{Concentration and sensitivity.}
Sparse-constraint candidate selection accounts for 74.98\% of the FTB-Core net gain. Its Base score is 32.50\%, above the uniform four-choice reference of 25\%; numeric bond-angle and signed-dihedral tasks have no comparable four-choice chance baseline. CubeAssembly accounts for 54.17\% of the SpatialViz net gain, with a 1.25\% Base score; 2DRotation starts at 0\%. Both are below four-choice uniform chance. These scores motivate checking output validity rather than assuming that all changes reflect reasoning.

Excluding sparse selection leaves a $+1.11\pp$ FTB-Core gain over 11,000 questions. Excluding CubeAssembly leaves $+3.00\pp$ on 1,100 SpatialViz questions; also excluding 2DRotation leaves $+2.84\pp$ on 1,020. Excluding both complete benchmarks leaves $+2.77\pp$ over the other eight General-10 datasets. These post hoc reaggregations concern Full, preserve original weights within each retained benchmark, and do not correct parser failures. 

\paragraph{Video tasks.}
VSI improves most on relative direction ($+4.92\pp$) and room-size estimation ($+3.80\pp$), while absolute distance ($-0.85\pp$) and route planning ($-1.89\pp$) decline. The remaining changes are appearance order $+0.65$, object counting $+0.79$, relative distance $+1.74$, and object-size estimation $+1.54\pp$.

% BEGIN INLINED OUTPUT-VALIDITY AUDIT: core-results/output-format-audit-20260915/appendix.tex
\subsection{Output validity, extraction failures, and answer-selection bias}
\label{app:output_validity}

\paragraph{Protocol and denominators.}
We audit all 12,000 FTB-Core and 1,180 SpatialViz predictions from Base and each
of the three canonical Full adapters (seeds 20260729, 20260803, and 20260804),
without regenerating answers or changing the scorer. Replaying the repository's
parsers exactly reproduces every archived prediction and correctness label.
Base outputs are identical across the three paired evaluations. Validity means
that the parser extracts a supported categorical label or a finite number;
a wrong answer or a numeric answer outside tolerance remains valid. The reference
answer is not used to decide validity. With $N$ total questions, $V$ valid outputs,
and $C$ correct answers, parse-failure rate is $(N-V)/N$, all-item accuracy is $C/N$,
and valid-output-only accuracy is $C/V$.

Conditioning each model on its own valid outputs changes the evaluated subset.
We therefore also report the accuracy difference on the identical IDs for which
both Base and the adapter produce valid outputs, separately for each seed before
averaging. All Full outputs in these two benchmarks are valid, so this common-valid
subset is also the fixed Base-valid subset. These are diagnostic conditional
comparisons, not counterfactual estimates of reasoning with output format held fixed.

\paragraph{FTB-Core.}
Base and every Full seed have 12,000/12,000 valid outputs. Valid-output-only and
all-item accuracy therefore coincide: 36.18\% for Base and 40.26\% for Full.
Sparse-constraint candidate selection likewise has 1,000/1,000 valid outputs in
both conditions and improves from 32.50\% to 69.17\%; all outputs for this task
re-encode to a single answer token. Its gain is not recovery from failed parsing.
However, Base selects A/B on 946/1,000 questions despite approximately balanced
reference labels, so answer-selection bias is a distinct possible contributor.
The task contributes 74.98\% of the FTB net gain; the remaining eleven tasks gain
$1.11\pp$ on average. These observations limit a claim of uniform spatial improvement.

% BEGIN INLINED OUTPUT-VALIDITY AUDIT: core-results/output-format-audit-20260915/ftb_validity.tex
\begin{table}[htbp]
\centering
\caption{FTB-Core parser audit. PF: parse-failure rate (percent of all rows); VO: accuracy conditional on a valid output (percent). Common-valid $\Delta$ uses the same IDs for Base and Full in each seed. Full values are means of three seed-wise rates, not pooled votes. Overall rates use question weighting.}
\label{tab:format_ftb_core}
\scriptsize
\setlength{\tabcolsep}{3pt}
\begin{tabular}{lrrrrrr}
\toprule
Task & $n$ & Base PF & Full PF & Base VO & Full VO & Common $\Delta$ \\
\midrule
all & 12000 & 0.00 & 0.00 & 36.18 & 40.26 & +4.08 \\
Bond angle & 1000 & 0.00 & 0.00 & 0.60 & 0.23 & -0.37 \\
Constraint audit & 1000 & 0.00 & 0.00 & 9.40 & 7.50 & -1.90 \\
Fragment assembly & 1000 & 0.00 & 0.00 & 28.30 & 30.10 & +1.80 \\
Nearest neighbor & 1000 & 0.00 & 0.00 & 41.50 & 50.47 & +8.97 \\
Template selection & 1000 & 0.00 & 0.00 & 25.80 & 26.97 & +1.17 \\
Rigid equivalence & 1000 & 0.00 & 0.00 & 47.20 & 48.20 & +1.00 \\
Signed dihedral & 1000 & 0.00 & 0.00 & 0.40 & 0.37 & -0.03 \\
Sparse-constraint selection & 1000 & 0.00 & 0.00 & 32.50 & 69.17 & +36.67 \\
Sphere collision & 1000 & 0.00 & 0.00 & 53.60 & 55.80 & +2.20 \\
Path clearance & 1000 & 0.00 & 0.00 & 49.30 & 48.70 & -0.60 \\
Chirality & 1000 & 0.00 & 0.00 & 49.40 & 49.37 & -0.03 \\
Contact evidence & 1000 & 0.00 & 0.00 & 96.20 & 96.23 & +0.03 \\
\bottomrule
\end{tabular}
\end{table}
% END INLINED OUTPUT-VALIDITY AUDIT: core-results/output-format-audit-20260915/ftb_validity.tex

\paragraph{SpatialViz.}
Base has 20 parse failures (1.69\%), compared with zero for each Full seed.
Valid-output-only accuracy is 27.07\% for Base ($314/1,160$) and 32.71\% for
Full (three-seed mean on 1,180 rows). On the same 1,160 Base-valid questions,
accuracy is 27.07\% versus 32.53\%, giving $+5.46\pp$; the seed-wise gains are
$5.60$, $5.09$, and $5.69\pp$. The 20 Base-invalid questions yield 9, 9, and 8
correct Full answers. Their mean contribution to the original all-item gain is
$0.7345\pp$, or 12.04\% of the $6.1017\pp$ net gain. The other 87.96\% is
the net improvement on questions already parseable for Base. This decomposition
does not establish that all recovered answers are caused by format learning,
or that all common-valid improvement is caused by better reasoning.

CubeAssembly (1.25\% versus 50.00\%) and 2DRotation (0.00\% versus 5.00\%) each
have 80/80 valid outputs for Base and every Full seed. Base selects D on 79/80
CubeAssembly and 74/80 2DRotation questions; none of their reference answers is D.
The poor scores therefore reflect valid wrong choices, not rejected answer strings.
CubeAssembly contributes 54.17\% of the SpatialViz net gain, but removing it still
leaves $+3.00\pp$. A conservative, target-independent sensitivity parser accepting
singleton parenthesized/lowercase letters and unambiguous explicit answer cues
recovers no additional correct SpatialViz answers. Output-format recovery, correction
of a response bias, and improved visual inference must not be conflated.

% BEGIN INLINED OUTPUT-VALIDITY AUDIT: core-results/output-format-audit-20260915/spatialviz_validity.tex
\begin{table}[htbp]
\centering
\caption{SpatialViz parser audit. PF: parse-failure rate (percent of all rows); VO: accuracy conditional on a valid output (percent). Common-valid $\Delta$ uses the same IDs for Base and Full in each seed. Full values are means of three seed-wise rates, not pooled votes. Overall rates use question weighting.}
\label{tab:format_spatialviz}
\scriptsize
\setlength{\tabcolsep}{3pt}
\begin{tabular}{lrrrrrr}
\toprule
Task & $n$ & Base PF & Full PF & Base VO & Full VO & Common $\Delta$ \\
\midrule
all & 1180 & 1.69 & 0.00 & 27.07 & 32.71 & +5.46 \\
2DRotation & 80 & 0.00 & 0.00 & 0.00 & 5.00 & +5.00 \\
3DRotation & 80 & 0.00 & 0.00 & 30.00 & 35.42 & +5.42 \\
3ViewProjection & 100 & 0.00 & 0.00 & 31.00 & 38.67 & +7.67 \\
ArrowMoving & 80 & 0.00 & 0.00 & 30.00 & 33.75 & +3.75 \\
BlockMoving & 80 & 0.00 & 0.00 & 31.25 & 31.25 & +0.00 \\
CrossSection & 120 & 0.83 & 0.00 & 14.29 & 11.39 & -3.64 \\
CubeAssembly & 80 & 0.00 & 0.00 & 1.25 & 50.00 & +48.75 \\
CubeCounting & 120 & 6.67 & 0.00 & 42.86 & 44.72 & +2.68 \\
CubeReconstruction & 120 & 0.00 & 0.00 & 30.00 & 30.56 & +0.56 \\
CubeUnfolding & 120 & 0.00 & 0.00 & 33.33 & 30.00 & -3.33 \\
MechanicalSystem & 80 & 13.75 & 0.00 & 50.72 & 52.92 & +3.38 \\
PaperFolding & 120 & 0.00 & 0.00 & 27.50 & 33.89 & +6.39 \\
\bottomrule
\end{tabular}
\end{table}
% END INLINED OUTPUT-VALIDITY AUDIT: core-results/output-format-audit-20260915/spatialviz_validity.tex

\paragraph{Length-limit diagnostics.}
The archived records contain decoded text but not generation token IDs or stopping
reasons, so exact truncation rates cannot be identified and are reported as unavailable,
not zero. We re-encode the decoded text with the local Qwen3.5-9B tokenizer, without
special tokens, as a length-limit diagnostic. All 20 Base-invalid SpatialViz strings
re-encode to the 128-token cap; they occur in MechanicalSystem (11), CubeCounting (8),
and CrossSection (1). No other SpatialViz output reaches that decoded-length threshold.
Neither Base nor Full FTB text reaches its 32-token cap. Re-encoding after special-token
removal is not an exact reconstruction of generation length or termination reason.

\paragraph{Robustness and scope.}
After omitting FTB-Core and SpatialViz entirely, the remaining eight General-10
benchmarks improve by $2.648$, $2.680$, and $2.981\pp$ across seeds (mean $2.770\pp$,
sample SD $0.184\pp$). This is a post-hoc sensitivity analysis with the canonical
GraphQA scorer. The audit addresses extraction failures in the two spatial
benchmarks; it does not isolate format effects throughout General-10. For example,
a valid alternative answer representation in a text benchmark can still fail its
scorer. The strongest supported interpretation is task-dependent behavioral transfer,
including improvements among already parseable spatial answers. Establishing a
format-independent spatial mechanism requires additional interventions such as
fixed-choice decoding and balanced option permutations.

\section{Complete Component Ablation Results}
\label{app:component_ablation}

\begin{table}[h]
\centering
\caption{Dataset-level component ablation results. Values are mean accuracies in
percent over three seeds; parentheses report absolute change from \base{} in
percentage points. The w/o FoldingCorpus arm removes FoldingCorpus CE, the w/o Geometry arm
removes the Geometry objective, and Fold2Reason-full combines FoldingCorpus CE and Geometry.
The historical w/o FoldingCorpus arm is geometry-dominant and includes the legacy retrieval
auxiliary.}
\scriptsize
\setlength{\tabcolsep}{2.6pt}
\begin{tabular*}{\linewidth}{@{\extracolsep{\fill}}ccccc@{}}
\toprule
\textbf{Benchmark} & \textbf{\base{}} & \textbf{w/o FoldingCorpus} &
\textbf{w/o Geometry} & \textbf{Fold2Reason-full} \\
\midrule
FTB-Core & 36.18 & 38.44 (+2.26) & 39.00 (+2.81) & \textbf{40.26 (+4.08)} \\
SpatialViz & 26.61 & 27.40 (+0.79) & 31.81 (+5.20) & \textbf{32.71 (+6.10)} \\
VSI & 58.53 & 58.58 (+0.05) & 59.08 (+0.54) & \textbf{60.16 (+1.63)} \\
GraphQA Easy & 65.14 & 67.34 (+2.20) & 68.72 (+3.58) & \textbf{69.70 (+4.56)} \\
GraphQA Hard & 32.65 & 32.63 (-0.02) & \textbf{44.44 (+11.79)} & 39.48 (+6.83) \\
BBH & 54.45 & 54.64 (+0.19) & 55.17 (+0.72) & \textbf{56.81 (+2.36)} \\
ChemBench & 66.76 & 66.98 (+0.22) & \textbf{67.55 (+0.79)} & 67.49 (+0.73) \\
ChemBench4K & 63.56 & 65.46 (+1.90) & 64.87 (+1.31) & \textbf{67.93 (+4.37)} \\
Lab-Bench & 37.21 & 36.55 (-0.66) & \textbf{38.33 (+1.12)} & 37.86 (+0.64) \\
SciBench & 9.83 & 9.66 (-0.17) & \textbf{11.26 (+1.44)} & 10.86 (+1.03) \\
\midrule
\textbf{General-10 macro} & \textbf{45.09} & 45.77 (+0.68) &
48.02 (+2.93) & \textbf{48.33 (+3.23)} \\
\bottomrule
\end{tabular*}
\end{table}

The component table reports the full dataset profile behind
Table~\ref{tab:components}. The w/o FoldingCorpus arm comes from the sealed component
evaluation, while Fold2Reason-full uses the canonical run. These shared-contract
comparisons isolate the behavioral and structural roles of the two loss paths; a full
factorial interaction analysis remains a separate experiment.

\subsection{Seed-level component comparisons}

FoldingCorpus supervision accounts for most of the broad transfer in each training
replicate. The mean Full-minus-w/o-Geometry increment is $0.303\pp$, with
individual-seed differences of $+0.575$, $+0.486$, and $-0.152\pp$.
Its t interval spans zero, whereas each arm's gain over Base remains positive
in all three seeds. This is a conditional loss comparison with the workspace retained; the direct Pure-LoRA comparison does not show a resolved aggregate advantage for the additional modules.

\begin{table}[H]
\centering
\caption{Component gains and the paired Geometry increment on General-10 (percentage points).}
\label{tab:ae_component_seed}
\scriptsize
\setlength{\tabcolsep}{3pt}
\begin{tabular*}{\linewidth}{@{\extracolsep{\fill}}lrrrrr@{}}
\toprule
Comparison & S1 & S2 & S3 & Mean $\pm$ SD & 95\% t interval \\
\midrule
w/o FoldingCorpus & +0.688 & +0.783 & +0.556 & $0.676 \pm 0.114$ & [+0.392, +0.959] \\
w/o Geometry & +2.611 & +2.629 & +3.551 & $2.931 \pm 0.538$ & [+1.595, +4.266] \\
Full RG & +3.186 & +3.115 & +3.400 & $3.234 \pm 0.148$ & [+2.866, +3.601] \\
Full $-$ w/o Geometry & +0.575 & +0.486 & -0.152 & $0.303 \pm 0.396$ & [-0.681, +1.287] \\
\bottomrule
\end{tabular*}
\vspace{3pt}
\parbox{0.98\linewidth}{\footnotesize The w/o FoldingCorpus arm is the historical geometry-dominant arm with its legacy retrieval auxiliary, as in the main component table.}
\end{table}

\section{Native Language-Head Folding Evaluation}
\label{app:native_folding}

The unadapted Qwen3.5-9B baseline receives the query sequence and available
MSA/template evidence, without reference coordinates, and generates one
C$\alpha$ coordinate row per residue through its native LM head. We use greedy
decoding with thinking disabled and a budget of $\max(2048,24L+1024)$ new tokens
for sequence length $L$. \method{} uses the existing epoch-3 checkpoints for
seeds 20260729, 20260803, and 20260804 with their workspace and frozen decoder.

We map finite coordinate rows to query positions by unique residue indices,
excluding duplicate and out-of-range indices. The primary mapping uses the
index rather than the echoed amino-acid character. Partial predictions are retained
without coordinate imputation. TM-score uses Kabsch alignment \citep{kabsch1976rotation} on available positions
and normalization by the full reference-valid target length. lDDT-C$\alpha$
retains all reference pairs within 15~\AA\ in its denominator; missing endpoints
receive no distance-preservation credit. Contact F1 counts missing true contacts
as false negatives, using an 8~\AA\ threshold and minimum separation of six positions
in the reference-valid residue ordering. These rules reproduce the existing scorer
for complete predictions.

All 334 proteins receive equal weight, including the three without usable
coordinates, which score zero. Native predictions cover 97.22\% of reference-valid
residues on average. Requiring matching amino-acid characters additionally gives
native scores of 2.05/2.22/1.05 for TM-score/lDDT-C$\alpha$/Contact F1
($\times100$), with the same 334-protein denominator.

\section{Structural Readout Audits}
\label{app:structural_audits}
Table~\ref{tab:structural_readout_complete} gives the complete FoldBench334 structural
means referenced in Section~\ref{sec:ablations}. Paired per-protein distributions,
per-protein maps, and fixed-quantile examples appear below; seed-level means are
available in \path{foldbench_audit.json} and \path{appendix_tables.json}.

\begin{table}[H]
\centering
\caption{Complete FoldBench334 structural readout comparison over three seeds. The
frozen-base row trains a decoder of the same architecture on frozen Qwen features;
the three adapted rows use the Phase-0 decoder as a fixed readout. The historical
w/o FoldingCorpus arm is geometry-dominant and includes a legacy retrieval auxiliary.
Lower C$\alpha$ MAE is better.}
\label{tab:structural_readout_complete}
\small
\setlength{\tabcolsep}{4.0pt}
\begin{tabular*}{\linewidth}{@{\extracolsep{\fill}}ccccc@{}}
\toprule
\textbf{Readout condition} & \textbf{TM-score $\uparrow$} &
\textbf{lDDT-C$\alpha$ $\uparrow$} & \textbf{Contact F1 $\uparrow$} &
\textbf{C$\alpha$ MAE (\AA) $\downarrow$} \\
\midrule
Frozen-base readout & \textbf{0.1742} & 0.2400 & 0.0664 & \textbf{10.753} \\
w/o Geometry & 0.1710 & 0.2435 & 0.0627 & 12.096 \\
w/o FoldingCorpus & 0.1685 & \textbf{0.2534} & \textbf{0.0696} & 12.262 \\
Fold2Reason-full & 0.1688 & 0.2532 & 0.0690 & 12.244 \\
\bottomrule
\end{tabular*}
\end{table}

\subsection{Per-protein structural effects}
\label{app:geometry_distribution}

\begin{figure}[H]
  \centering
  \includegraphics[width=0.999\linewidth]{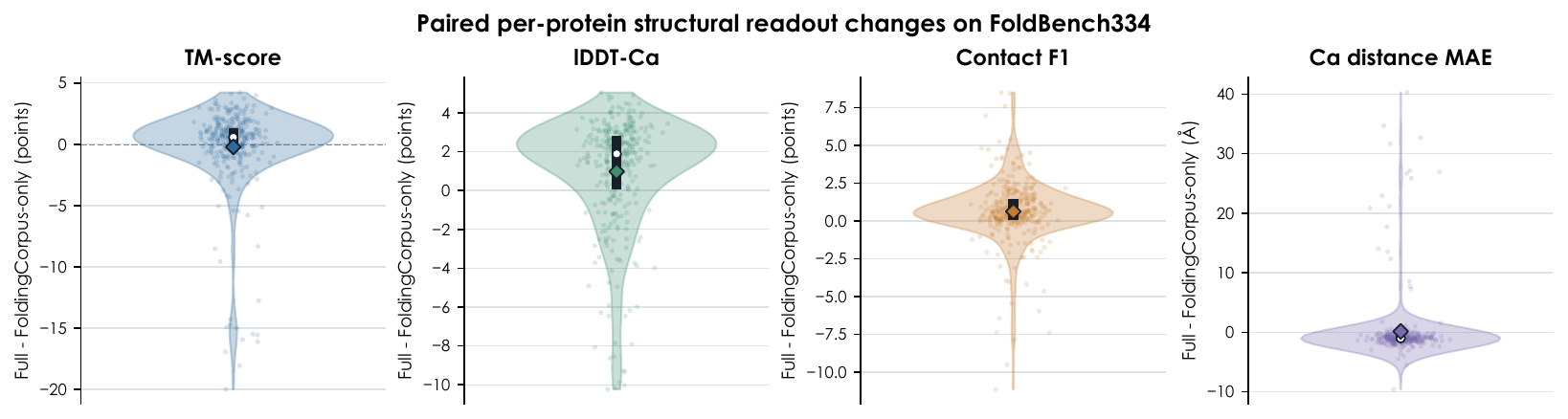}
  \caption{\textbf{Geometry changes local readouts more consistently than global
  topology.} Each point is one FoldBench334 protein after averaging that protein over
  three training seeds; distributions show Fold2Reason-full minus FoldingCorpus-only.
  Thick bars span the interquartile range, white circles mark medians, and diamonds
  mark means. Positive values favor Full except for C$\alpha$ MAE, where negative is
  better. Full improves per-protein lDDT-C$\alpha$ and Contact F1 for 75.4\% and
  78.7\% of proteins, respectively, while TM-score and MAE contain influential
  tails.}
  \label{fig:geometry_readout_distributions}
\end{figure}

Figure~\ref{fig:geometry_readout_distributions} resolves the cohort-level means
into paired changes for individual proteins. Each comparison averages the three
training seeds for a protein before subtracting FoldingCorpus-only from the full
model. Here FoldingCorpus-only means the workspace-equipped w/o Geometry arm, not Pure LoRA. The distributions show the consistency of local readout improvements and
the variation in global-structure metrics.

\begin{figure}[H]
  \centering
  \includegraphics[width=0.96\linewidth]{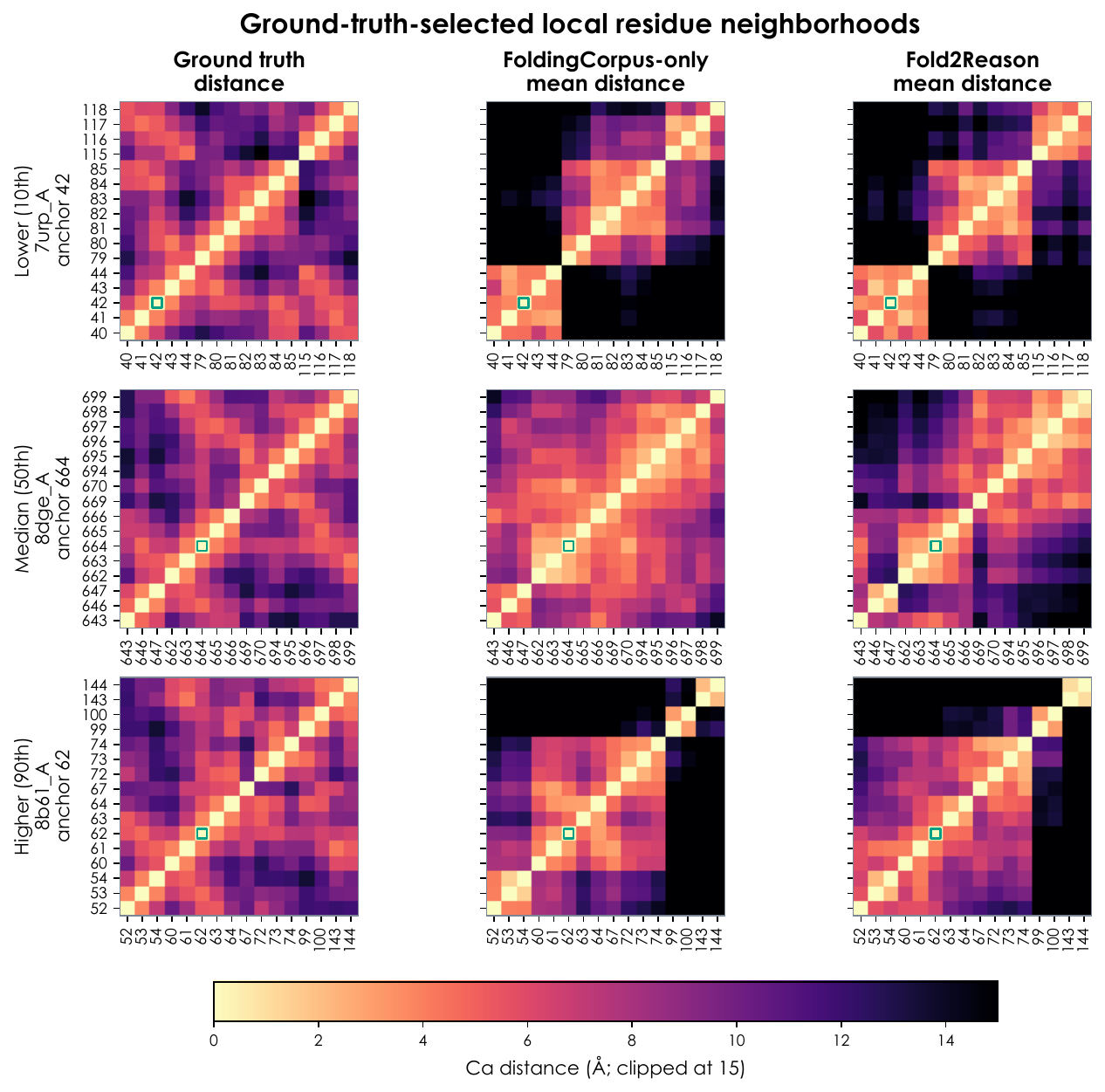}
  \caption{\textbf{Local distance-map audit for the preselected proteins of
  Table~\ref{tab:structural_quantile_cases}.}
  For each protein, the anchor is the residue with the most ground-truth contacts; the
  displayed neighborhood contains that anchor and its 15 nearest ground-truth
  C$\alpha$ neighbors. Ground truth alone determines the selection. Full changes the
  local distance MAE from 6.311 to 6.178~\AA{} in the lower case, from 2.620 to
  2.821~\AA{} in the median case, and from 5.982 to 5.533~\AA{} in the higher case.
  The middle case worsens while the lower and higher cases improve, matching the
  heterogeneous Geometry effect in the aggregate analysis. Green squares mark the
  anchor residue.}
  \label{fig:local_structural_neighborhoods}
\end{figure}

Table~\ref{tab:structural_quantile_cases} consolidates the two predeclared qualitative
audits: Panel A reports the three proteins in Figure~\ref{fig:structural_contact_distance_maps},
while Panel B retains the lower, median, and higher Contact-F1-change cases used in
Figure~\ref{fig:local_structural_neighborhoods}. The latter include one protein whose
contact F1 decreases.

\begin{table}[h]
\centering
\caption{Preselected FoldBench334 cases for qualitative structural audit. Panel A
matches Figure~\ref{fig:structural_contact_distance_maps}: the three proteins have the
highest Full mean TM-score among all 334 targets. Panel B matches
Figure~\ref{fig:local_structural_neighborhoods}: proteins are selected at the 10th,
50th, and 90th percentiles of the per-protein Contact F1 change. Selection uses the
three-seed mean, with protein ID breaking ties, and precedes visual inspection. Each
metric cell reports FC-only $\rightarrow$ Full, followed by Full minus FC-only in
parentheses; lower distance MAE is better.}
\label{tab:structural_quantile_cases}
\scriptsize
\setlength{\tabcolsep}{1.8pt}
\renewcommand{\arraystretch}{1.18}
\begin{tabular*}{\linewidth}{@{\extracolsep{\fill}}llrcccc@{}}
\toprule
\textbf{Selection} & \textbf{Protein} & \textbf{$L$} &
\shortstack[c]{\textbf{TM-score}\\\textbf{FC-only $\rightarrow$ Full}} &
\shortstack[c]{\textbf{lDDT-C$\alpha$}\\\textbf{FC-only $\rightarrow$ Full}} &
\shortstack[c]{\textbf{Contact F1}\\\textbf{FC-only $\rightarrow$ Full}} &
\shortstack[c]{\textbf{Distance MAE (\AA)}\\\textbf{FC-only $\rightarrow$ Full}} \\
\midrule
\multicolumn{7}{l}{\textit{Panel A: proteins displayed in Figure~\ref{fig:structural_contact_distance_maps}}} \\
Rank 1 & 8wt3\_A & 134 & \shortstack{0.2438 $\rightarrow$ 0.2742\\(+0.0304)} & \shortstack{0.2767 $\rightarrow$ 0.3034\\(+0.0267)} & \shortstack{0.1146 $\rightarrow$ 0.1169\\(+0.0023)} & \shortstack{6.0422 $\rightarrow$ 5.3820\\(-0.6602)} \\
Rank 2 & 8qjp\_A & 250 & \shortstack{0.2255 $\rightarrow$ 0.2586\\(+0.0331)} & \shortstack{0.2462 $\rightarrow$ 0.2629\\(+0.0167)} & \shortstack{0.0377 $\rightarrow$ 0.0552\\(+0.0175)} & \shortstack{10.8590 $\rightarrow$ 9.1586\\(-1.7004)} \\
Rank 3 & 7xg9\_A & 286 & \shortstack{0.2310 $\rightarrow$ 0.2568\\(+0.0258)} & \shortstack{0.2607 $\rightarrow$ 0.2834\\(+0.0227)} & \shortstack{0.0435 $\rightarrow$ 0.0493\\(+0.0058)} & \shortstack{12.2372 $\rightarrow$ 10.8577\\(-1.3794)} \\
\midrule
\multicolumn{7}{l}{\textit{Panel B: Contact-F1-change quantiles displayed in Figure~\ref{fig:local_structural_neighborhoods}}} \\
Lower (10th) & 7urp\_A & 159 & \shortstack{0.1917 $\rightarrow$ 0.2035\\(+0.0118)} & \shortstack{0.2125 $\rightarrow$ 0.2275\\(+0.0150)} & \shortstack{0.0707 $\rightarrow$ 0.0595\\(-0.0112)} & \shortstack{9.3934 $\rightarrow$ 8.0564\\(-1.3371)} \\
Median (50th) & 8dge\_A & 707 & \shortstack{0.1506 $\rightarrow$ 0.1616\\(+0.0110)} & \shortstack{0.2169 $\rightarrow$ 0.2480\\(+0.0312)} & \shortstack{0.0249 $\rightarrow$ 0.0307\\(+0.0058)} & \shortstack{27.6176 $\rightarrow$ 26.2962\\(-1.3214)} \\
Higher (90th) & 8b61\_A & 197 & \shortstack{0.2324 $\rightarrow$ 0.2364\\(+0.0040)} & \shortstack{0.2294 $\rightarrow$ 0.2608\\(+0.0314)} & \shortstack{0.0876 $\rightarrow$ 0.1126\\(+0.0249)} & \shortstack{9.5911 $\rightarrow$ 8.3614\\(-1.2297)} \\
\bottomrule
\end{tabular*}
\end{table}

\clearpage

The intervention rows are an appendix-only workspace audit. On
FoldBench334, Fold2Reason-full obtains TM-score 0.1688, lDDT-C$\alpha$ 0.2532,
contact F1 0.0690, and C$\alpha$ distance MAE 12.24~\AA. Replacing a matched
workspace with one from another protein reduces TM-score by 0.01043, showing that the
geometry decoder uses protein-specific residue
information.

\subsection{Why the structural cohort contains 334 proteins}
\label{app:foldbench_cohort}

FoldBench334 contains the entire monomer-protein target list represented by the
local FoldBench release manifest, whose source is
\texttt{targets/monomer\_protein.csv}. The manifest has 334 distinct target IDs.
All 334 appear in the matched structural-evaluation records for every seed of
Full RG, w/o Geometry, w/o FoldingCorpus, and the frozen-base decoder control.
Thus, the cohort size follows the monomer-protein target list rather than a
performance-based subsample. Other FoldBench task categories are outside this
monomer readout evaluation.

The manifest sequences range from 28 to 1,414 residues. The structural evaluation
uses the residue representation in the cached input and coordinate records,
which can be shorter than the manifest sequence: the minimum evaluated length
is 26 residues and the mean is 261.10, compared with 261.43 in the manifest.
We report both distributions to distinguish target coverage from residue-level
preprocessing. Protein IDs, rather than sequence length alone, define the paired
comparisons.

\begin{table}[H]
\centering
\caption{FoldBench334 cohort size and residue-count distribution. Manifest sequence length and the residue count in the structural evaluation cache are reported separately.}
\label{tab:ae_fold_cohort}
\scriptsize
\setlength{\tabcolsep}{3pt}
\begin{tabular*}{\linewidth}{@{\extracolsep{\fill}}lrrrrrrr@{}}
\toprule
Representation & N & Min & Q1 & Median & Q3 & Max & Mean \\
\midrule
Manifest sequence & 334 & 28.0 & 140.2 & 225.5 & 344.8 & 1414.0 & 261.43 \\
Evaluated residues & 334 & 26.0 & 140.2 & 225.5 & 344.8 & 1414.0 & 261.10 \\
\bottomrule
\end{tabular*}
\end{table}

\subsection{Structural readout scope}
Complete target coverage and data isolation are separate. The three adapted arms use the same Phase-0 fixed reader; the frozen-base arm trains a separate decoder of the same architecture. Full improves local readouts relative to w/o Geometry but has lower TM-score and worse distance MAE. FoldBench homology, deposition-date, and fold-level isolation remain incomplete, and the independent scaling audit found template-source and alignment overlaps (Appendix~\ref{app:scaling_dynamics}). These measurements remain diagnostics under the stated conditions.

\section{Data Scaling and Checkpoint Dynamics}
\label{app:scaling_dynamics}

The scaling study uses nested subsets of the same frozen training pool and the three
canonical seeds. Every protein contributes all 12 FoldingCorpus labels, and every scale is
trained for three epochs. All endpoints use the complete
held-out FoldingCorpus split, FoldBench334, and General-10; no endpoint was selected using
evaluation performance. This is an independent scaling run, which accounts for its
$3.31\pp$ 1,000-protein endpoint differing slightly from the canonical $3.23\pp$
main experiment. The complete study covers seven training-set sizes from 50 to
4,000 proteins under the same three-epoch schedule. The 4,000-protein pool uses a
mean-pLDDT/fraction-high-confidence gate of 70/0.70, versus 80/0.80 for the
2,000-protein pool; this comparison changes both sample count and quality support.

\begin{table}[h]
\centering
\caption{Fixed-epochs data-scaling endpoints. Values are three-seed means from the
independent scaling run. General-10 values are changes from the base model in
percentage points.}
\label{tab:data_scaling_endpoints}
\small
\setlength{\tabcolsep}{3.0pt}
\begin{tabular*}{\linewidth}{@{\extracolsep{\fill}}ccccccc@{}}
\toprule
\textbf{Proteins} & \shortstack{\textbf{FoldingCorpus}\\\textbf{targets}} & \textbf{Steps} &
\shortstack{\textbf{FoldingCorpus}\\\textbf{macro}} & \textbf{lDDT-C$\alpha$} & \textbf{Contact F1} &
\textbf{General-10 $\Delta$} \\
\midrule
50 & 600 & 21 & 0.478 & 0.244 & 0.0666 & 0.49 \\
100 & 1,200 & 39 & 0.486 & 0.244 & 0.0671 & 0.61 \\
250 & 3,000 & 96 & 0.485 & 0.246 & 0.0666 & 1.37 \\
500 & 6,000 & 189 & 0.492 & 0.247 & 0.0675 & 2.92 \\
1,000 & 12,000 & 375 & 0.501 & 0.252 & 0.0694 & 3.31 \\
2,000$^\dagger$ & 24,000 & 750 & 0.546 & 0.252 & 0.0739 & 3.70 \\
4,000$^\dagger$ & 48,000 & 1,500 & 0.418 & 0.252 & 0.0768 & 3.02 \\
\bottomrule
\end{tabular*}

\vspace{2pt}
\parbox{0.98\linewidth}{\footnotesize $^\dagger$Fresh-run extension beyond the original 50--1,000-protein curve. The
4,000-protein FoldingCorpus mean has high seed variance because one seed exhibits
answer-token calibration collapse; the predeclared greedy metric is retained.}
\end{table}

\begin{figure}[t]
  \centering
  \includegraphics[width=\linewidth]{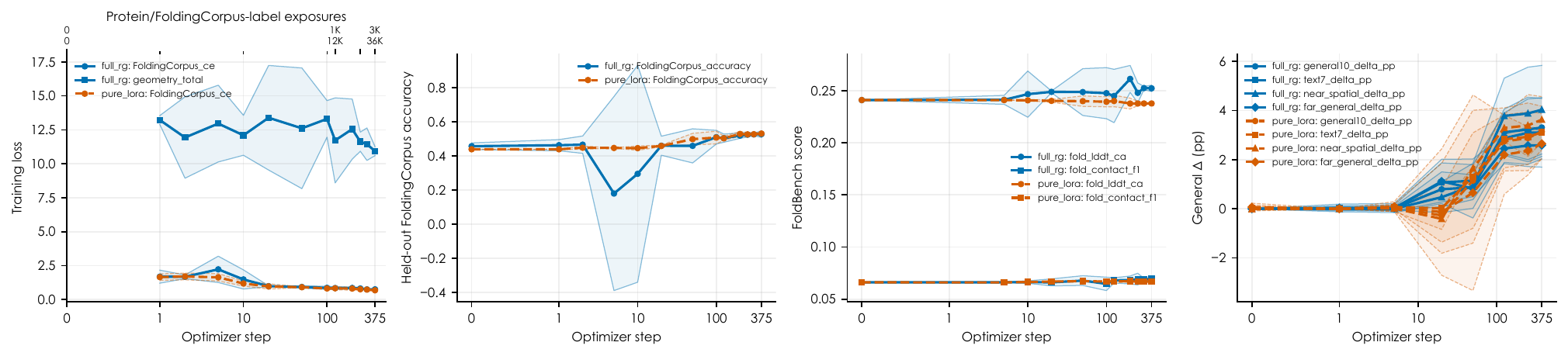}
  \caption{\textbf{Training and transfer dynamics at pre-registered checkpoints.}
  Columns show training losses, held-out FoldingCorpus answer accuracy, FoldBench
  structural readouts, and General-10 changes. Legend entries carry the run
  identifiers \texttt{full\_rg} for Fold2Reason-full (blue) and \texttt{pure\_lora}
  for FoldingCorpus-only Pure-LoRA (orange). The upper axis reports cumulative
  protein and answer-label exposures. All benchmark evaluations were run offline
  after training and were not used for checkpoint selection.}
  \label{fig:checkpoint_dynamics}
\end{figure}

Figure~\ref{fig:checkpoint_dynamics} tracks the pre-registered checkpoints for Full RG
and FoldingCorpus-only Pure-LoRA. At step 5, the three Full-RG General-10 changes are
$-0.02$, $+0.03$, and $+0.04\pp$; by step 375 they reach $+3.15$, $+2.94$, and
$+3.85\pp$. Pure-LoRA follows the same broad timing pattern, moving from
$+0.03/+0.05/+0.08\pp$ at step 5 to $+2.80/+3.25/+3.37\pp$ at step 375. Under the
pre-registered trajectory rules, every seed in both arms is classified as continuous
growth, with none classified as an instant plateau, transient peak, or decoupling.
The gain therefore develops over optimization rather than appearing as an immediate
prompt-format response; gradual format adaptation remains an alternative explanation, and this timing pattern is not specific to the Geometry loss.

\paragraph{Observed overlaps and audit scope.}
Six scaling-training inputs contain template-contact constraints originating from FoldBench targets: 8ec3, 8ey3, 8g64, 8t9n, 8uds, and 9etn. A Smith--Waterman audit \citep{smith1981waterman} at $\geq30\%$ identity and $\geq80\%$ bidirectional coverage found 12 train--dev, 14 train--test, three train--FoldBench, three dev--test, and one dev--FoldBench matches. PDB IDs were not model inputs, but this does not remove template-derived evidence. Checkpoints and evaluation sets were not changed after discovery, and no cleaned rerun is claimed. FoldBench curves are therefore readout diagnostics, not strictly template- or remote-homology-disjoint generalization. The identified matches concern protein-side data rather than General-10 items; they do not constitute a complete audit of base-model pretraining contamination.

\subsection{Evaluation contracts and complete scale-by-seed results}
\label{app:scaling_numeric_details}

The independent scaling study has its own frozen base generations and archived
scoring implementation. Table~\ref{tab:ae_protocol_bases} records these alongside
the canonical endpoint base to make the comparison unit explicit. In particular,
the scaling study uses VSI's official eight-task macro, while the canonical
endpoint study uses the mean score over all questions. The other differing base
scores reflect the respective archived evaluation runs. All points within the
scaling and bandwidth analyses use the same scaling base, and all reported
changes are computed within their respective evaluation contract.

\begin{table}[H]
\centering
\caption{The two frozen base-evaluation contracts used by the canonical endpoint study and the independent scaling study. Each reported gain uses the base from its own column.}
\label{tab:ae_protocol_bases}
\scriptsize
\setlength{\tabcolsep}{3pt}
\begin{tabular*}{\linewidth}{@{\extracolsep{\fill}}lrrr@{}}
\toprule
Dataset & N & Canonical base & Scaling base \\
\midrule
FTB-Core & 12000 & 36.18 & 36.23 \\
SpatialViz & 1180 & 26.61 & 26.61 \\
VSI & 5130 & 58.53 & 56.93 \\
GraphQA Easy & 21600 & 65.14 & 67.71 \\
GraphQA Hard & 21600 & 32.65 & 33.39 \\
BBH & 6511 & 54.45 & 54.00 \\
ChemBench & 2148 & 66.76 & 66.53 \\
ChemBench4K & 4009 & 63.56 & 61.29 \\
Lab-Bench & 1967 & 37.21 & 36.25 \\
SciBench & 580 & 9.83 & 9.14 \\
\bottomrule
\end{tabular*}
\vspace{3pt}
\parbox{0.98\linewidth}{\footnotesize VSI uses mean-question scoring in the canonical endpoint study and the official eight-task macro in the archived scaling study. Text/FTB base generations also belong to their respective runs.}
\end{table}

Table~\ref{tab:ae_scaling_seed} reports all 21 fixed-three-epoch endpoints. The
4K run with seed 20260803 has frozen-test FoldingCorpus accuracy 0.1283, compared with
0.5683 for seed 20260729. Its General-10 change remains positive at
$3.139\pp$. Retaining this run exposes the answer-calibration instability
behind the 4K FoldingCorpus mean and the difference between source-task accuracy
and external transfer.

\begin{table}[H]
\centering
\caption{All fixed-three-epoch data-scaling endpoints by seed. FoldingCorpus and structure metrics are fractions; General-10 changes are percentage points.}
\label{tab:ae_scaling_seed}
\scriptsize
\setlength{\tabcolsep}{3pt}
\begin{tabular*}{\linewidth}{@{\extracolsep{\fill}}lrrrrrr@{}}
\toprule
Proteins & Seed & Steps & FC acc. & lDDT & Contact F1 & General-10 $\Delta$ \\
\midrule
50 & 20260729 & 21 & 0.4608 & 0.2384 & 0.0650 & +0.406 \\
50 & 20260803 & 21 & 0.4875 & 0.2458 & 0.0667 & +0.686 \\
50 & 20260804 & 21 & 0.4858 & 0.2474 & 0.0680 & +0.392 \\
100 & 20260729 & 39 & 0.4758 & 0.2432 & 0.0660 & +0.232 \\
100 & 20260803 & 39 & 0.4975 & 0.2432 & 0.0670 & +0.593 \\
100 & 20260804 & 39 & 0.4850 & 0.2469 & 0.0682 & +1.013 \\
250 & 20260729 & 96 & 0.5208 & 0.2489 & 0.0658 & +1.457 \\
250 & 20260803 & 96 & 0.4392 & 0.2459 & 0.0667 & +1.487 \\
250 & 20260804 & 96 & 0.4958 & 0.2430 & 0.0675 & +1.165 \\
500 & 20260729 & 189 & 0.4892 & 0.2477 & 0.0677 & +2.971 \\
500 & 20260803 & 189 & 0.4842 & 0.2477 & 0.0680 & +3.043 \\
500 & 20260804 & 189 & 0.5017 & 0.2464 & 0.0666 & +2.743 \\
1000 & 20260729 & 375 & 0.4867 & 0.2516 & 0.0692 & +3.145 \\
1000 & 20260803 & 375 & 0.5175 & 0.2537 & 0.0700 & +2.941 \\
1000 & 20260804 & 375 & 0.5000 & 0.2515 & 0.0690 & +3.848 \\
2000$^\dagger$ & 20260729 & 750 & 0.5508 & 0.2530 & 0.0748 & +3.763 \\
2000$^\dagger$ & 20260803 & 750 & 0.5367 & 0.2499 & 0.0735 & +3.468 \\
2000$^\dagger$ & 20260804 & 750 & 0.5492 & 0.2525 & 0.0736 & +3.876 \\
4000$^\dagger$ & 20260729 & 1500 & 0.5683 & 0.2511 & 0.0754 & +2.780 \\
4000$^\dagger$ & 20260803 & 1500 & 0.1283 & 0.2525 & 0.0767 & +3.139 \\
4000$^\dagger$ & 20260804 & 1500 & 0.5567 & 0.2531 & 0.0782 & +3.152 \\
\bottomrule
\end{tabular*}
\vspace{3pt}
\parbox{0.98\linewidth}{\footnotesize $^\dagger$Fresh-run extension beyond the original 50--1,000-protein curve. The 4K pool also changes the structure-confidence support. Values preserve the scoring protocol of the independent scaling study described in this section.}
\end{table}

\subsection{Which datasets account for data-scaling gains?}

The dataset profiles show different responses to increasing protein coverage.
GraphQA Hard and SpatialViz contribute substantial gains toward 2K proteins,
while ChemBench4K peaks earlier in the displayed range. ChemBench and Lab-Bench
have small negative mean changes at 4K. The macro therefore summarizes a mixture
of strengthening, saturating, and declining task responses. All seven scales
and all ten datasets are included in the table and heatmap.

\begin{table}[H]
\centering
\caption{Dataset-level changes along the complete fixed-three-epoch curve (percentage points). The final row reports the macro mean and training-seed SD.}
\label{tab:ae_scaling_dataset}
\scriptsize
\setlength{\tabcolsep}{3pt}
\begin{tabular*}{\linewidth}{@{\extracolsep{\fill}}lrrrrrrr@{}}
\toprule
Dataset & 50 & 100 & 250 & 500 & 1K & 2K$^\dagger$ & 4K$^\dagger$ \\
\midrule
FTB-Core & -0.13 & -0.44 & +0.50 & +4.26 & +4.26 & +5.00 & +4.56 \\
SpatialViz & +0.76 & +0.65 & +1.53 & +5.17 & +5.93 & +7.12 & +6.92 \\
VSI & +0.12 & -0.08 & +0.33 & +1.17 & +1.13 & +1.82 & +1.58 \\
GraphQA Easy & +1.36 & +2.13 & +1.77 & +2.58 & +2.49 & +2.76 & +1.47 \\
GraphQA Hard & +0.78 & +0.56 & +1.94 & +3.90 & +6.35 & +8.67 & +8.46 \\
BBH & +0.57 & +0.74 & +2.37 & +3.30 & +3.77 & +3.68 & +2.90 \\
ChemBench & -0.19 & -0.85 & -0.62 & +0.47 & +1.30 & +0.34 & -0.12 \\
ChemBench4K & +0.58 & +2.06 & +3.60 & +6.03 & +5.61 & +4.78 & +2.74 \\
Lab-Bench & +0.29 & +0.15 & +0.85 & +0.47 & +0.37 & +0.44 & -0.39 \\
SciBench & +0.80 & +1.21 & +1.44 & +1.84 & +1.90 & +2.41 & +2.13 \\
General-10 & $0.49 \pm 0.17$ & $0.61 \pm 0.39$ & $1.37 \pm 0.18$ & $2.92 \pm 0.16$ & $3.31 \pm 0.48$ & $3.70 \pm 0.21$ & $3.02 \pm 0.21$ \\
\bottomrule
\end{tabular*}
\end{table}

\begin{figure}[H]
  \centering
  \includegraphics[width=\linewidth]{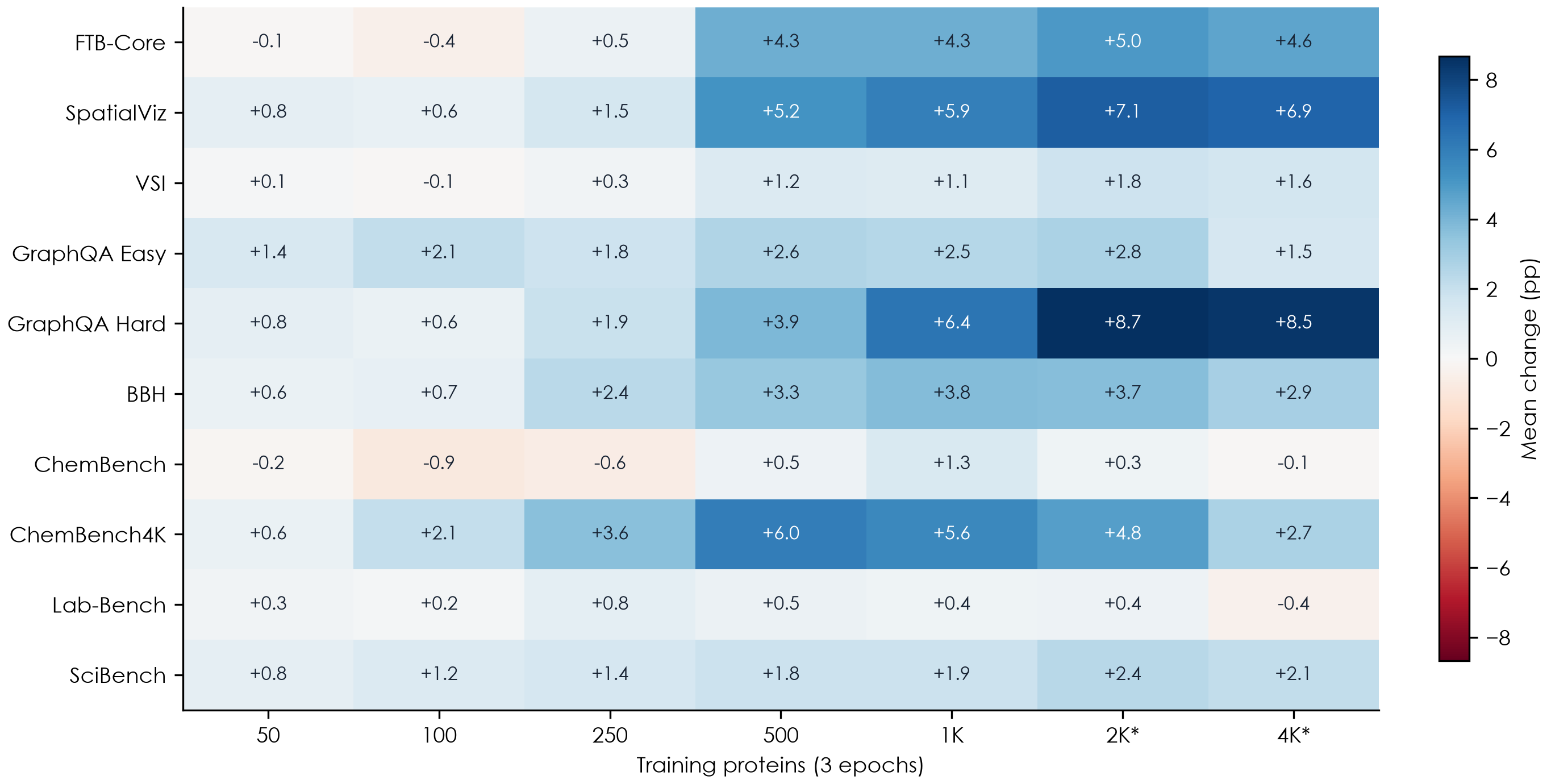}
  \caption{\textbf{Dataset-level transfer along the fixed-three-epoch curve.}
  Cells show the three-seed mean change from the scaling study's frozen base in
  percentage points. A common diverging color scale is centered at zero.
  Asterisks mark the fresh 2K and 4K extension runs; the 4K pool also changes
  the confidence-quality support described above.}
  \label{fig:appendix_scaling_dataset}
\end{figure}

\subsection{FoldingCorpus-label density at fixed protein coverage}
\label{app:label_bandwidth}

The label-density experiment holds the training set at 1,000 proteins and the
schedule at 375 optimizer steps, then uses 3, 6, or 12 FoldingCorpus labels per protein.
The q=12 condition is the independent 1K scaling endpoint. The mean General-10
changes are $3.391$, $2.951$, and $3.311\pp$, respectively. Increasing label
density therefore produces no monotonic transfer increase in this range.
The mean within-seed slope against $\log_2(q)$ is $-0.040\pp$ per doubling,
with a 95\% t interval of $[-0.814,0.734]$. The three-seed interval permits both
positive and negative density effects, so it neither supports a positive dose--response
relationship nor establishes that label density has no effect. In particular, these
results do not support the claim that asking more structural questions about the same
proteins improves transfer over the tested range. Label count is an operational measure
of supervision density, not a direct measure of independent structural information.
Redundancy among labels or saturation by three labels could explain a flat response,
but neither explanation is established by this experiment. The protein-count curve
also increases data breadth and training compute together and cannot resolve this
mechanism. We therefore interpret the combined evidence as behavioral transfer from
protein-derived supervision, without attributing that transfer to increasing label
density or claiming that these experiments identify reusable reasoning computations.

\begin{table}[H]
\centering
\caption{FoldingCorpus-label density at fixed 1,000 proteins and 375 optimizer steps. S1--S3 and the first mean report General-10 changes (pp); the remaining metrics are fractions.}
\label{tab:ae_bandwidth}
\scriptsize
\setlength{\tabcolsep}{3pt}
\begin{tabular*}{\linewidth}{@{\extracolsep{\fill}}lrrrrrrr@{}}
\toprule
Labels/protein & S1 & S2 & S3 & $\Delta \pm$ SD & FC acc. & lDDT & Contact F1 \\
\midrule
3 & +2.971 & +3.731 & +3.472 & $3.391 \pm 0.386$ & $0.4942 \pm 0.0175$ & $0.2544 \pm 0.0020$ & $0.0690 \pm 0.0007$ \\
6 & +3.145 & +2.441 & +3.268 & $2.951 \pm 0.446$ & $0.4994 \pm 0.0047$ & $0.2542 \pm 0.0028$ & $0.0693 \pm 0.0003$ \\
12 & +3.145 & +2.941 & +3.848 & $3.311 \pm 0.476$ & $0.5014 \pm 0.0155$ & $0.2523 \pm 0.0012$ & $0.0694 \pm 0.0005$ \\
\bottomrule
\end{tabular*}
\end{table}

\begin{figure}[H]
  \centering
  \includegraphics[width=\linewidth]{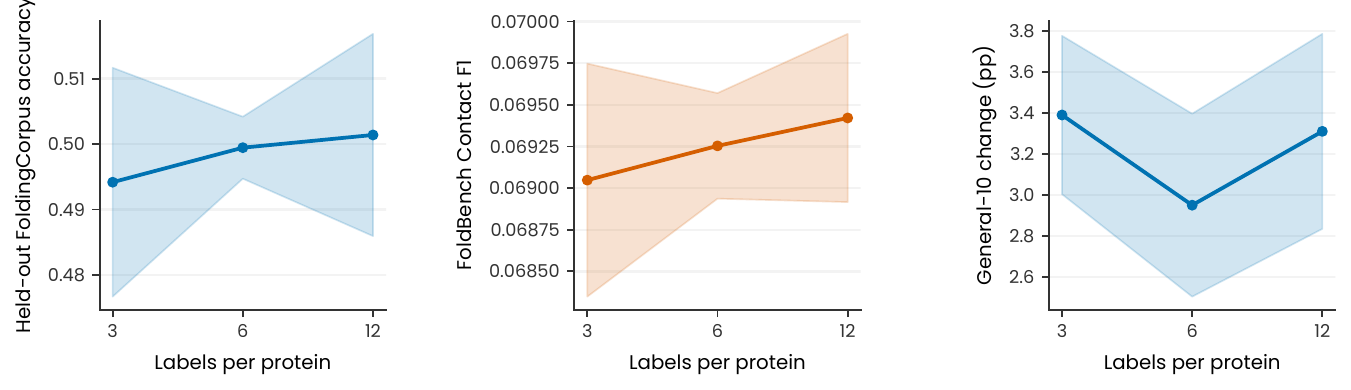}
  \caption{\textbf{FoldingCorpus-label density with 1,000 proteins and 375 steps.}
  Solid lines show means over three training seeds. Shaded regions extend from
  mean minus one sample SD to mean plus one sample SD; the thin boundary lines
  mark those limits. FoldingCorpus accuracy uses the frozen test split. FoldBench
  uses all 334 targets, and external transfer uses the scaling evaluation contract.}
  \label{fig:appendix_label_bandwidth}
\end{figure}

\subsection{Numerical checkpoint trajectories}
\label{app:checkpoint_numeric}
Figure~\ref{fig:checkpoint_dynamics} retains the complete recorded trajectories. FoldingCorpus accuracy there uses the 100-protein development split; scaling and label-density endpoint tables use the separate 100-protein frozen test split. Full's General-10 gain is $0.016\pp$ at step 5, $3.116\pp$ at step 125, and $3.311\pp$ at step 375; Pure LoRA reaches $3.140\pp$ at step 375 in this independent study. The canonical Full/Pure endpoints are a separate experiment. Numerical losses, scheduled gaps, structural metrics, and every seed trajectory remain in \path{scaling_evidence.json} and \path{appendix_tables.json}. The time course alone does not distinguish reasoning acquisition from gradual answer-format adaptation.

\section{Broader Impact}
\label{app:broader_impact}

This work studies transfer from public protein structures into general model behavior.
Potential benefits include more data-efficient structured supervision and clearer
audits of what scientific post-training changes. The main risks are benchmark
contamination and misuse of structural readouts as folding predictions. Sequence
overlap audits, seed-level variation, and separate behavioral and decodability metrics
make these risks visible. The released models and documentation will identify the
geometry output as a training diagnostic and reserve biological interpretation for
validated structure-prediction systems.

\section{Machine-Readable Evidence Files}
\label{app:evidence_manifest}

The supplementary source directory
\texttt{core-results/appendix-evidence-20260906/} contains the numerical evidence
behind the tables and figures of this paper. The table builder reads the archived
results, checks paired identifiers and aggregate reconstruction, and exports the
following files.

\begin{table}[H]
\centering
\caption{Machine-readable evidence included with the appendix source package.}
\label{tab:ae_artifacts}
\scriptsize
\begin{tabularx}{\linewidth}{@{}lX@{}}
\toprule
File & Contents \\
\midrule
\texttt{endpoint\_seed\_scores.json} & All ten dataset scores by seed for Fold2Reason-full, the component arms, the model families, and the source-control aggregates. \\
\texttt{model\_training\_contracts.json} & Saved LoRA modules, parameter counts, steps, seed, and hardware world size for the fifteen model-family runs. \\
\texttt{task\_breakdown.json} & Complete FTB split/family/task, SpatialViz category/task, and VSI question-group scores. \\
\texttt{foldbench\_audit.json} & The 334 target IDs, sequence and evaluation length distributions, and structural metrics by arm and seed. \\
\texttt{scaling\_evidence.json} & Fixed-three-epoch endpoints, label-density results, and numerical checkpoint trajectories. \\
\texttt{scaling\_evaluation\_manifest.json} & Archived checkpoint and split assignments and expected evaluation counts for the included original scaling runs. \\
\texttt{scaling\_recorded\_hashes.json} & Archived adapter-state, evaluation-ID, and base-model hashes, including completion records for the 2K and 4K extensions. \\
\texttt{appendix\_source\_manifest.csv} & Paths, SHA-256 digests, and byte counts of the source files used by the builder. \\
\texttt{appendix\_tables.json} & Every generated table cell and its caption. \\
\bottomrule
\end{tabularx}
\end{table}

Every reported General-10 endpoint covers 76{,}725 questions across the ten datasets.
The structural audit checks all 334 protein IDs in every arm and seed. FTB subgroup
counts reconstruct the 12{,}000-example total, and the SpatialViz and VSI decompositions
use the same prediction IDs across Base and all three adapted arms. The
machine-readable bundle preserves run-specific base and scoring contracts, the
development versus frozen-test FoldingCorpus splits, and the distinction between the
canonical and the independent scaling checkpoints. Rebuilding the tables from scratch
requires the referenced experiment files; inspecting the exported scores, table cells,
and figures requires only the supplementary source package.

\end{document}